\documentclass[acmsmall]{acmart}
\usepackage{algorithm}
\usepackage{algorithmic}
\usepackage{multirow}
\usepackage{pifont}
\usepackage{forest}
\usepackage{tikz}
\usepackage{graphicx}
\usepackage{wrapfig}
\definecolor{hidden-draw}{RGB}{20,68,106}
\definecolor{LemonChiffon}{RGB}{255,250,205}
\definecolor{Gold}{RGB}{255,215,0}
\definecolor{LightBlue}{RGB}{173,216,230}
\definecolor{SteelBlue}{RGB}{70,130,180}
\definecolor{LightCyan}{RGB}{225,255,255}
\pdfoutput=1
\AtBeginDocument{%
  \providecommand\BibTeX{{%
    \normalfont B\kern-0.5em{\scshape i\kern-0.25em b}\kern-0.8em\TeX}}}

\begin{document}

\title{A Survey of Models for Cognitive Diagnosis: New Developments and Future Directions}



\author{Fei Wang}
\author{Weibo Gao}
\author{Qi Liu}
\author{Jiatong Li}
\author{Guanhao Zhao}
\author{Zheng Zhang}
\author{Zhenya Huang}
\affiliation{%
  \institution{State Key Laboratory of Cognitive Intelligence, University of Science and Technology of China}
  \streetaddress{Jinzhai Road No.96}
  \city{Hefei}
  \country{China}
}
\author{Mengxiao Zhu}
\affiliation{%
  \institution{Anhui Province Key Laboratory of Science Education and Communication, University of Science and Technology of China}
  \streetaddress{Jinzhai Road No.96}
  \city{Hefei}
  \country{China}
}
\author{Shijin Wang}
\affiliation{%
  \institution{iFLYTEK AI Research (Central China)}
  \city{Wuhan}
  \country{China}
}
\author{Wei Tong}
\affiliation{%
  \institution{National Educational Examinations Authority}
  \city{Beijing}
  \country{China}
}
\author{Enhong Chen}
\affiliation{%
  \institution{State Key Laboratory of Cognitive Intelligence, University of Science and Technology of China}
  \streetaddress{Jinzhai Road No.96}
  \city{Hefei}
  \country{China}
}
\authorsaddresses{
  Authors' addresses: F. Wang, W. Gao, J. Li, G. Zhao, Z. Zhang, Q. Liu, Z. Huang and Enhong Chen, State Key Laboratory of Cognitive Intelligence, University of Science and Technology of China, Hefei, China; emails: \{wf314159, weibogao, satosasara, ghzhao0223, zhangzheng\}@mail.ustc.edu.cn and \{qiliuql, huangzhy, cheneh\}@ustc.edu.cn; M. Zhu, Anhui Province Key Laboratory of Science Education and Communication, University of Science and Technology of China; email: mxzhu@ustc.edu.cn; S. Wang, iFLYTEK AI Research (Central China), Wuhan, China; email: sjwang3@iflytek.com; W. Tong, National Educational Examinations Authority; email: tongw@mail.neea.edu.cn.
}

\renewcommand{\shortauthors}{Fei Wang, et al.}

\begin{abstract}
  Cognitive diagnosis has been developed for decades as an effective measurement tool to evaluate human cognitive status such as ability level and knowledge mastery. It has been applied to a wide range of fields including education, sport, psychological diagnosis, etc. By providing better awareness of cognitive status, it can serve as the basis for personalized services such as well-designed medical treatment, teaching strategy and vocational training. This paper aims to provide a survey of current models for cognitive diagnosis, with more attention on new developments using machine learning-based methods. By comparing the model structures, parameter estimation algorithms, model evaluation methods and applications, we provide a relatively comprehensive review of the recent trends in cognitive diagnosis models. Further, we discuss future directions that are worthy of exploration. In addition, we release two Python libraries: EduData for easy access to some relevant public datasets we have collected, and EduCDM that implements popular CDMs to facilitate both applications and research purposes.
\end{abstract}

\ccsdesc[500]{Applied computing~Computer-assisted instruction}
\ccsdesc[500]{Applied computing~E-learning}
\ccsdesc[500]{Social and professional topics~Computing education}
\ccsdesc[100]{Social and professional topics~Computing profession}
\ccsdesc[100]{Social and professional topics~User characteristics}

\keywords{Cognitive diagnosis, item response theory, cognitive diagnosis model, intelligent education, deep learning, survey}


\maketitle

\section{Introduction}
Measurement is an integral part of modern science as well as of engineering, commerce, and daily life \cite{tal2015measurement}. Through measurement, we learn quantitatively about the things around us and even human beings. Cognitive diagnosis, as a representative, aims to measure the cognitive status of individuals, especially the ability levels such as knowledge structures and processing skills, so as to provide information about their cognitive strengths and weaknesses \cite{leighton2007cognitive}. For example, through cognitive diagnosis, we can learn about whether a student has mastered specific knowledge concepts \cite{leighton2007cognitive} or whether a patient is mentally healthy \cite{de2018analysis}. Therefore, cognitive diagnosis provides informative results for test developers and test takers, as well as helps with personalized support such as training course planning for employers and learning resource recommendations for students.

\begin{figure}
    \centering
    \includegraphics[width=0.9\textwidth]{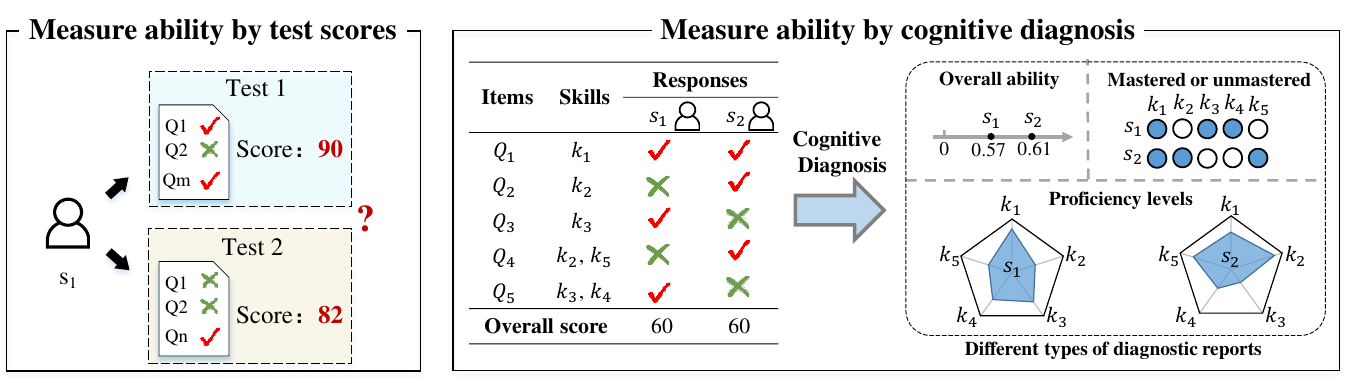}
    \caption{Comparison between ability measurement by test scores and by cognitive diagnosis.}
    \label{fig:CD_irf}
    \vspace{-0.1in}
\end{figure}

Unlike conventional physical measurement objects such as length and weight, a person's ability level is a psychological characteristic not directly observable. Therefore, the fundamental idea of measuring human ability is by conducting tests and inferring examinees' ability through their performance. Francis Galton is thought as the first to apply statistical methods to the study of human differences and inheritance of intelligence, and proposed the first personality test~\cite{rust2014modern}. Using the scores obtained in a test is a straightforward way to represent a person's overall ability level, and is widely adopted in IQ tests, teaching, etc. However, although test scores reflect examinees' ability level to some extent, they are not the ability level themselves. For instance, as demonstrated in the left part of Fig. \ref{fig:CD_irf}, a person may get different scores on two tests at the same time, while neither of them can absolutely represent the correct ability level.
By contrast, cognitive diagnosis infers the ability level hidden in the responses. The complete cognitive diagnosis procedure (especially traditional cognitive diagnosis) requires multiple steps, including preparatory work such as deciding the measurement goal, arranging the knowledge structures, and constructing tests. A simplified procedure is illustrated in the right part of Fig. \ref{fig:CD_irf} which includes: 1) Test construction: rigorous questionnaires or test items (e.g., $Q_1 \sim Q_5$) can be constructed for response collection in fields such as education and psychotherapy \cite{henson2005test}. The relation between items and relevant attributes (i.e., skills or knowledge concepts) are usually provided by experts. 2) Response data collection: the responses here mostly refer to binary 0/1 values indicating the results of examinees' answers (e.g., incorrect/correct) or discrete scores obtained on the items (e.g., 5 points out of 8). In some situations, responses are not limited to question answering, for example, the outcome of adversarial games~\cite{gu2021neuralac} and law cases~\cite{an2021lawyerpan}. 3) Cognitive diagnosis model (CDM) designing: well-designed CDMs are an important guarantee of valid diagnostic results. 4) Psychological factor estimation: based on the collected response data, the psychological factors (e.g., the ability parameters) within CDMs are estimated. 4) Diagnosis feedback: the diagnosed ability levels are then fed back to the examinees. The feedback can be different depending on the CDMs, such as overall ability (\S\ref{sec:ability_level}), whether or not mastered certain attributes (\S\ref{sec:cognition_level}), and proficiency levels of certain attributes (\S\ref{sec:ml_deep}).

\begin{wrapfigure}{r}{0cm}
    \centering
    \includegraphics[width=0.5\textwidth]{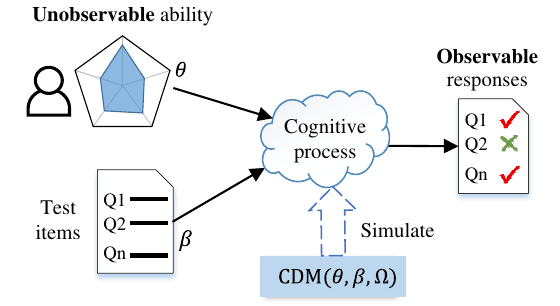}
    \caption{The essence of CDM.}
    \label{fig:CDM_essence}
    \vspace{-0.1in}
\end{wrapfigure}
Cognitive diagnosis models are essential for cognitive diagnosis, aiming to infer the \textbf{unobservable} ability levels from the \textbf{observable} responses to test items. Essentially, most existing CDMs are/contain simulations of examinees' cognitive processes. Specifically, as illustrated in Fig. \ref{fig:CDM_essence}, when answering the test items, examinees go through a cognitive process that handles the items with their knowledge status and then provide their responses to the items. The responses depend on multiple factors, including the characteristics of both items (e.g., difficulty~\cite{Reckase2009}, relevant knowledge concept, guessing~\cite{de2009dina}) and examinees (e.g., ability, gaming behavior~\cite{wu2017knowledge}). Therefore, the central problem of cognitive diagnosis is to model the relation between the examinees' ability levels and their observable behaviors such as responses to test items \cite{frederiksen2012test}. This simulation can be formulated as:
\begin{equation*}
    \displaywidth=\parshapelength\numexpr\prevgraf+2\relax
    Pr(response) = f(\theta, \beta, \Omega),
\end{equation*}
where $\theta, \beta$ are the parameters indicating the examinees' abilities and items' features, respectively. $\Omega$ represents the possible parameters required by the CDM itself (empty for some models).

Originating from \textbf{psychometrics}, the models for measuring human abilities have been developed for decades. The proposal of item response theory (IRT) can be traced back to the 1950s by Fredrick Lord \cite{lord1952theory}. IRT is a general framework for specifying mathematical functions that describe the interactions of persons and test items, where unidimensional parameters were adopted to describe the abilities of the persons. However, as suggested by researchers such as Glaser \cite{glaser1981future} and Mislevy \cite{mislevy2012foundations}, IRT and its previous test theories (e.g., classical test theory \cite{devellis2006classical}) only measure the \textbf{macro} ability of individuals. Psychology was suggested to be combined with psychometrics in order to model the \textbf{micro} knowledge structure and cognitive processing of persons during the assessments so that the diagnostic results can be more instructional. The term \textbf{\textit{cognitive diagnosis model (CDM)}} is originally adopted to denote such models\footnote{The usage of the terminology in academia has not achieved a complete consensus. Strictly speaking, IRT-like models are mostly regarded as the predecessor of CDM. For convenience and without significant violation of the original definition, in this paper, we use \textbf{cognitive diagnosis model (CDM)} to denote all the models for cognitive diagnosis.}. The proposal and usage of Q-matrix was a significant milestone of CDM \cite{tatsuoka1983rule}. Subsequently, representative models such as AHM \cite{leighton2004attribute}, DINA \cite{de2009dina} and NIDA \cite{junker2001cognitive} were proposed based on different assumptions to simulate the knowledge structures or cognitive processes, and each examinee is classified into a mastery pattern representing his/her mastery of each specific skill. These models fall into the cognition level paradigm \cite{mislevy2012foundations}.

In recent years, some researchers have been rethinking the issue of cognitive diagnosis from the perspective of \textbf{machine learning} and have proposed novel solutions \cite{liu2021towards}. Collaborative filtering and matrix factorization methods were adopted to model learners' ability and to predict learners' test performance~\cite{toscher2010collaborative,thai2010recommender,thai2015multi}. Gierl et al. \cite{gierl2008using} proposed a neural network-based ability classifier trained with the data generated by a pre-trained attribute hierarchical model. More recently, Wang et al. recognized the limitations of expert-designed interaction functions and proposed a new data-driven cognitive diagnosis framework called NeuralCD \cite{wang2020neural}. Deep learning-based models incorporated with the theories/hypotheses from psychometrics have the advantage of better fitting ability of the sophisticated cognitive process, as well as promising interpretability. Since then, such \textbf{deep learning-based paradigm} gradually becomes a new tendency and has been attracting increasing attention~\cite{ma2022knowledge,wang2022neuralcd,wang2021using}. In addition to the usage of various deep learning technologies, this paradigm takes the naturally saved data of learners' daily behaviors into consideration, breaking the limitation of intentional tests. Furthermore, more types of data were further explored as supplements for cognitive diagnosis, including the knowledge structures \cite{gao2021rcd,li2022hiercdf}, the text of test items \cite{cheng2019dirt,wang2022neuralcd}, and the context features of learners \cite{zhou2021modeling}.

In terms of application areas, although cognitive diagnosis used to mainly focus on diagnosing students' knowledge mastery and patients' psychological disorders in the past, nowadays cognitive diagnosis has been more widely applied in various areas. For example, traditional IRT was also adopted to model the ability of workers in data crowdsourcing to improve the accuracy of truth inference \cite{whitehill2009whose,bachrach2012grade}. Gu et al. \cite{gu2021neuralac} considered the effect of cooperation and competition among game players' abilities to predict the outcomes of matches. An et al. \cite{an2021lawyerpan} diagnosed the proficiency of trial lawyers in different legal fields by proposing a lawyer proficiency assessment network.

\subsection{Goal and Contributions}
The goal of this survey is to summarize the past studies of cognitive diagnosis models, allowing newcomers to have a comprehensive understanding of cognitive diagnosis. At the same time, we will emphasize recent developments in cognitive diagnosis, and provide our understanding about new trends in the research.

The contributions are summarized as follows:
\begin{itemize}
    \item We review the development stages of cognitive diagnosis models in the past, including the recent new trends in research.
    \item We provide a more complete and comprehensive review of the recent deep learning-based cognitive diagnosis models.
    \item We demonstrate the usefulness of cognitive diagnosis, including its downstream applications and its usage in various areas.
    \item We discuss the limitations of current cognitive diagnosis models and potential future research directions.
\end{itemize}

\begin{figure}[tb]
    \centering
    \begin{forest}
        for tree={
        scale=0.84,
        grow'=east,
        text centered,
        parent anchor=east,
        child anchor=west,
        anchor=west,
        rectangle,
		draw=hidden-draw,
		rounded corners,
        edge path={
                \noexpand\path [draw, \forestoption{edge}]
                (!u.parent anchor) -- +(5pt,0) |- (.child anchor)\forestoption{edge label};
            },
        },
        where level=1{text width=11.7em}{},
        where level=2{text width=10em}{},
        where level=3{text width=14.7em}{},
        [Paper\\ Structure, draw, align=center  
        [Cognitive Diagnosis Models, draw, calign=first
        [Overview of CDMs \S\ref{sec:overview}]
        [Psychometrics-based\\ Models \S\ref{sec:traditional_CD}, fill=yellow, align=center
        [Ability Level Paradigm \S\ref{sec:ability_level}, fill=LemonChiffon]
        [Cognition Level Paradigm \S\ref{sec:cognition_level}, fill=Gold]
        ]
        [Machine Learning-based\\ Models \S\ref{sec:ML_CD}, fill=cyan, align=center
        [Non-deep Learning-based Models \S\ref{sec:ml_nondeep}, fill=LightCyan]
        [Deep Learning-based Models \S\ref{sec:ml_deep}, fill=SteelBlue!50]
        ]
        ]
        [Parameter Estimation \S\ref{sec:estimation}, draw]
        [Model Evaluation \S\ref{sec:evaluation}, draw]
        [Application \S\ref{sec:application}, draw]
        [Datasets and Tools \S\ref{sec:datasetTool}, draw]
        [Future Research Directions \S\ref{sec:future}, text width=13em, draw]
        [Conclusion \S\ref{sec:conclusion}, draw]
        ]
    \end{forest}
    \caption{Scope and structure of the survey.}
    \label{fig:paper_structure}
\end{figure}
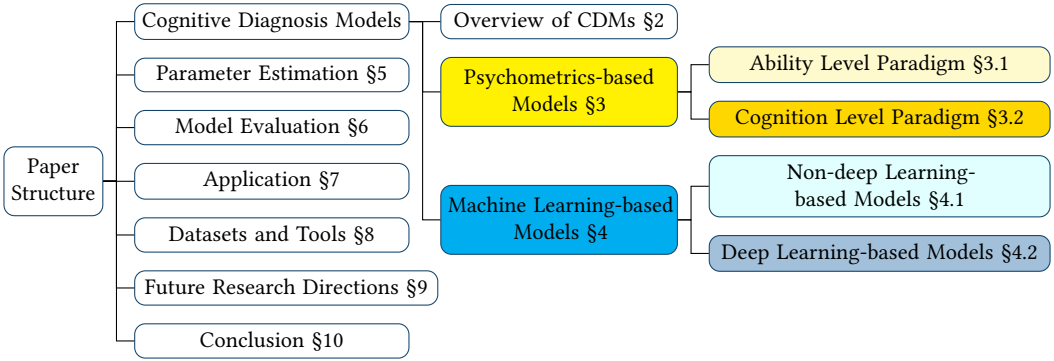

\subsection{Paper Structure}
In the rest of the paper, we review the research on cognitive diagnosis from the aspects of model structure, parameter estimation, evaluation, applications, datasets and future directions, as depicted in Fig \ref{fig:paper_structure}. Specifically, we firstly give an overview of CDMs in Section \ref{sec:overview}, and then introduce the details of existing models by classifying them into two main classes, i.e., psychometrics-based models (Section \ref{sec:traditional_CD}) and machine learning-based models (Section \ref{sec:ML_CD}). In Section \ref{sec:estimation}, we summarized the most commonly used algorithms for estimating the parameters in different types of CDMs. In Section \ref{sec:evaluation}, we summarize the evaluation of CDMs from multiple aspects. Applications based on cognitive diagnosis are summarized in Section \ref{sec:application}. Finally, we discuss the limitations of current research and potential research directions for future studies in Section \ref{sec:future}. The paper is concluded in Section \ref{sec:conclusion}.

\section{The Overview of Cognitive Diagnosis Models} \label{sec:overview}
\subsection{Preliminary} \label{sec:preliminary}
Suppose there are examinees $\mathcal{S}=\{S_1, \dots, S_I\}$, test items $\mathcal{Q}=\{Q_1, \dots, Q_J\}$ and $K$ item attributes (i.e., skills or knowledge concepts). The responses are denoted as $R=\{R_i, i=1, \dots, I\}$, where $R_i$ denotes the responses of $S_i$. $R_i=\{(S_i, Q_j, r_{ij}), S_i \in \mathcal{S}, Q_j \in \mathcal{Q}\}$ denotes $S_i$'s responses and $r_{ij}$ is the response result of $S_i$ on $Q_j$. Usually there is an expert-labeled Q-matrix $Q=\{q_{jk}\}^{J \times K}$, where $q_{jk}$=1 (or 0) indicating that item $Q_j$ requires (or does not require) the mastery of attribute $k$ to answer it correctly. Sometimes there are extra multifaceted information available, such as item content and examinees' background, which we denote as $X$. 
The problem of cognitive diagnosis can be generally defined as follows:

\textbf{Problem Definition.} With the input $R$, $Q$ and possible $X$, the goal of cognitive diagnosis is to output examinees' ability levels $\theta_i (i=1, \dots, I)$, where $\theta_i$ is either unidimensional or multidimensional indicating the overall ability levels or the mastery levels of each attribute.

The practicability of cognitive diagnosis is based on several basic assumptions.

\paragraph{Assumption 1} \textbf{Constant ability.} The cognitive status of concern, i.e., ability level, remains unchanged during the process of answering the test items.

It is reasonable to assume that a person's ability level does not change in a short time (e.g., during a standard test), during which the person's ability level can be measured based on the responses to test items. Here lies the big difference between cognitive diagnosis and knowledge tracing, of which the latter also attracted wide attention in recent years. Knowledge tracing focuses on modeling the changing patterns of online learners' knowledge states (either explainable or unexplainable), which highly relies on sequential modeling methods such as hidden Markov chains and Recurrent Neural Networks. The cognitive process is usually neglected in the knowledge tracing models, and predicting learners' future performance is the most adopted task. By contrast, cognitive diagnosis aims to measure learners' ability level within a certain period of time. It mines the response data of learners, models the cognitive process of answering items, and provides the values of learners' ability levels within a certain metric space. 

\paragraph{Assumption 2} \textbf{Constant item characteristics.} The characteristics of a test item remain constant over all of the testing situations where it is used \cite{Reckase2009}. 

Some statistics of the item such as the correct rate can be influenced by the examinees. However, the characteristics of a test item, such as the difficulty, discrimination, and relevant knowledge concepts, reflect the essential features of the item and should not change. This type of stability contributes to the fairness of test items for all examinees, and suggests that test items can be represented by fixed parameter values that reflect these characteristics.

\paragraph{Assumption 3} \textbf{Monotonicity.} The probability of a correct response to the test item increases, or at least does not decrease, as the locations of examinees increase on any of the coordinate dimensions \cite{Reckase2009}. 

Most cognitive diagnosis models adopt the monotonicity assumption in their modeling of cognitive processes, especially IRT-based and MIRT-based models. This assumption suggests that a better performance should result from a higher ability level, which is in accordance with common intuition or experience.

The above assumptions are the most adopted ones in cognitive diagnosis models. There are some model-specific assumptions in the research, leading to different model structures. For example, DINA \cite{de2009dina} and NIDA \cite{junker2001cognitive} models make detailed and different assumptions about how examinees' mastery of knowledge concepts decides their responses, and the Rule Space Method \cite{tatsuoka1983rule} assumes a hierarchical structure of knowledge concepts. We will provide some introduction in \S\ref{sec:traditional_CD} and \S\ref{sec:ML_CD}.

In the following content of the paper, we will use $\theta$ to represent examinees' ability; use $\beta$ to represent the features of test items, such as difficulty and discrimination. The features of test items could be used differently, and we will provide explanations when introducing specific CDMs. In addition, we will use $\Omega$ to represent all the parameters contained in a CDM itself besides the two types of parameters mentioned above. $\Omega$ only exists in the ML-based CDMs (\S\ref{sec:ML_CD}).

\subsection{A Brief Review of Cognitive Diagnosis Model Development}
Without cognitive diagnosis, the most widely adopted method to evaluate a learner's ability is through their scores obtained in tests. Classical Test Theory (CTT) \cite{devellis2006classical} was proposed to eliminate the errors existing in the scores. However, the score is the observed reflection of ability with the influence of factors such as question attributes and other psychological characteristics. To extract the actual ability hidden in the observations, cognitive diagnosis models sprouted from Psychometrics and have undergone decades of study. The development of cognitive diagnosis can be summarized from the aspects of both model structures and data characteristics. 
\begin{figure}
	\centering
	\includegraphics[width=\textwidth]{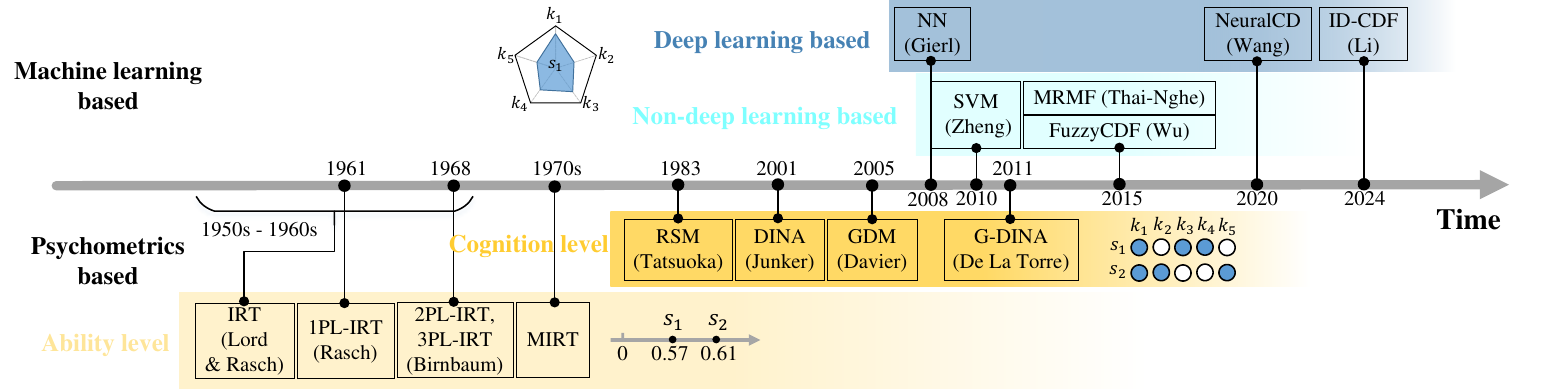}
	\caption{Representative models in the development of cognitive diagnosis model structures.}
	\label{fig:CDM_overview}
\end{figure}

\subsubsection{The development of model structures.}

Basically, the development of cognitive diagnosis models can be divided into two stages, i.e., psychometrics-based models and machine learning-based models. Fig. \ref{fig:CDM_overview} illustrates some of the representative works and the times they were proposed. Each stage of the development can be further divided into two sub-stages as follows:

\paragraph{Psychometrics-based models.} At the first stage of cognitive diagnosis development, ability measurement were based on psychometrics. Early research works were summarized as the \textbf{ability level paradigm} \cite{mislevy2012foundations}, as they used unidimensional or multidimensional latent vectors to represent examinees' overall ability levels. Representative methods include IRT and multidimensional IRT (MIRT). According to~\cite{Reckase2009}, its popularity is generally attributed to the work of Fredrick Lord and Georg Rasch starting in the 1950s and 1960s. Both IRT and MIRT have undergone lots of development and there is a large class of implementations. \S\ref{sec:ability_level} will provide a more detailed review of representative models. With the demand of measuring fine-grained ability, i.e., mastery of knowledge concepts or skills, the \textbf{cognition level paradigm} was proposed to improve diagnostic performance. The proposal of Q-matrix and its usage is a hallmark of these methods \cite{tatsuoka1983rule}. Representative works include RSM~\cite{tatsuoka1983rule}, DINA~\cite{junker2001cognitive}, GDM~\cite{vondavier2005gdm}, G-DINA~\cite{de2011generalized}, etc. Section \ref{sec:cognition_level} will provide more summary about such models.

\paragraph{Machine learning-based models.} Later in 2000s, there came increasingly more studies using machine learning (ML) models for cognitive diagnosis, such as clustering algorithms \cite{chiu2009cluster,guo2020spectral}, support vector machine~\cite{zheng2010application,liu2018application}, matrix factorization~\cite{toscher2010collaborative,thai2010recommender,thai2015multi} and fuzzy set \cite{wu2015cognitive}. \S\ref{sec:ml_nondeep} gives a review of relevant research using machine learning models for cognitive diagnosis before deep learning-based methods become popular. Although Gierl et al.~\cite{gierl2008using} made early attempts to use artificial neural networks as classifiers for cognitive diagnosis, the popularity of deep learning (DL)-based CDMs is mostly attributed to the work of Wang et al. \cite{wang2020neural} which preliminarily validated the superiority of using data-driven deep learning methods in cognitive diagnosis and inspired numerous research works \cite{gao2021rcd,gao2022deep,song2023deep,ma2022knowledge}. More recently, Li et al.~\cite{li2024towards} put emphasis on inductive diagnosis and proposed an encoder-decoder-like CDM. We will provide a comprehensive summary and comparison in \S\ref{sec:ML_CD}.

\subsubsection{The changes of exploited data.} 
\begin{figure}
	\centering
	\includegraphics[width=\textwidth]{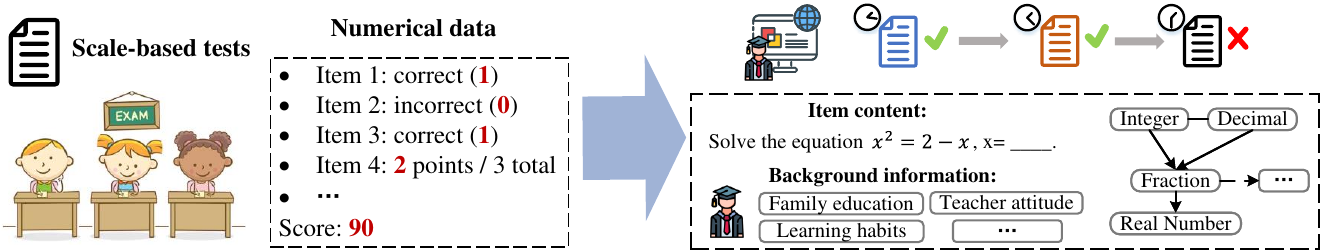}
	\caption{The changes of data types exploited by cognitive diagnosis.}
	\label{fig:CDM_data_evolve}
\end{figure}

As shown in the left part of Fig. \ref{fig:CDM_data_evolve}, in the early research works, the response data for diagnosing examinees' ability is collected from scale-based tests, where scales (e.g., questionnaires, test papers) are constructed and tests are intentionally organized. Only numerical data, i.e., correct, incorrect, and scores, is utilized in early psychometrics-based methods, such as IRT, MIRT, and DINA. After that, some psychometrics-based models leverage simple hierarchical structures among a small number of knowledge concepts by either using them to help with the defining of Q-matrix or explicitly modeling the hierarchical structures, such as AHM. Overall, the data types that can be utilized in pure psychometrics-based models are limited due to the simplicity of model structures.

With the usage of machine learning, researchers have been making attempts to leverage more types of data that contain relevant information for cognitive diagnosis. Especially after NeuralCD \cite{wang2020neural} validated the superiority of using deep learning methods in cognitive diagnosis, including better fitting ability and extensibility without losing explainability, following studies started to explore diverse types of data, including test item contents, examinees' background information and sophisticated graph-structured data (the right part of Fig. \ref{fig:CDM_data_evolve}). Moreover, the response data in consideration is not limited to scale-based tests. The popularity of online learning systems provides more opportunities of accessing learners' daily behavioral data as well as diverse data types. Therefore, cognitive diagnosis can be conducted even without organizing tests in an interruptive way if we regard learners' ability as constant during a short time period. Some researchers addressed the data sparsity problem within learners' response logs \cite{wang2022neuralcd, yao2023exploiting}, and considered supplementary data such as response time and hints \cite{wu2017knowledge,thai2015multi,zhan2018cognitive,zhan2022cognitive}. More detailed behaviors such as keystrokes and eye tracking can be available, however, they still need further exploration.

\section{Psychometrics-based Cognitive Diagnosis Models}  \label{sec:traditional_CD}
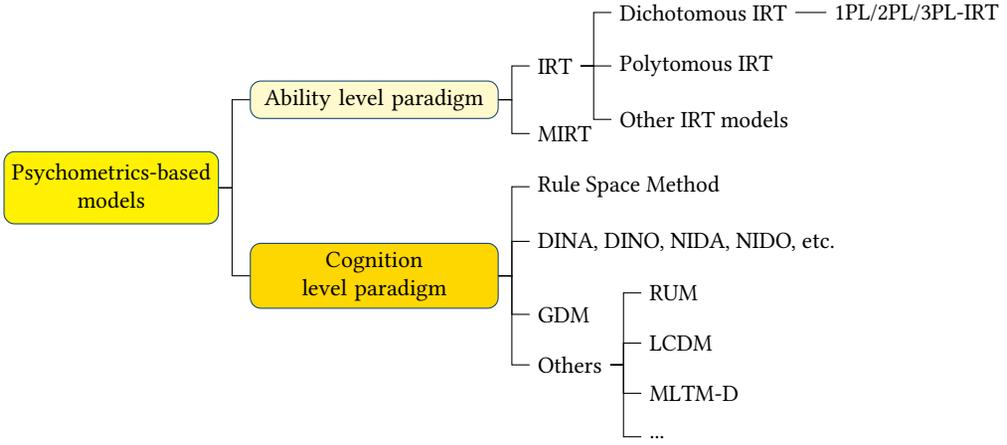
\begin{figure}[t]
    \centering
    \begin{forest}
        for tree={
            scale=0.88,
            grow'=east,
            text centered,
            rectangle,
            rounded corners,
            parent anchor=east,
            child anchor=west,
            anchor=west,   
            edge path={  
                \noexpand\path [draw, \forestoption{edge}]  
                (!u.parent anchor) -- +(5pt,0) |- (.child anchor)\forestoption{edge label};  
            },
        },
        where level=1{text width=10em}{},
        [Psychometrics-based\\ models, draw=hidden-draw, align=center, fill=yellow 
        [Ability level paradigm, draw=hidden-draw, fill=LemonChiffon
            [IRT
                    [Dichotomous IRT
                            [1PL/2PL/3PL-IRT]]
                    [Polytomous IRT]
                    [Other IRT models]
            ]
            [MIRT]
        ]
        [Cognition level paradigm, draw=hidden-draw, fill=Gold
            [Rule Space Method]
            [{DINA, DINO, NIDA, NIDO, etc.}]
            [GDM]
            [Others
                    [RUM]
                    [LCDM]
                    [MLTM-D]
                    [...]
            ]
        ]
        ]
    \end{forest}
    \caption{A taxonomy of psychometrics-based cognitive diagnosis models.}
    \label{fig:traditional-cd}
\end{figure}

Traditional cognitive diagnosis models (CDMs) utilize statistical methods based on psychometrics to model examinees' cognitive status. These models can be categorized into two paradigms, i.e., the \textbf{ability level paradigm} and the \textbf{cognition level paradigm} (Fig. \ref{fig:traditional-cd}).

\subsection{Ability Level Paradigm} \label{sec:ability_level}
\par In the ability level paradigm, researchers focus on the estimation of the \textit{overall ability} of examinees reflected on tests. Traditional models following the ability level paradigm usually model examinees' overall ability levels by low-dimensional $\theta$, which can be jointly estimated with low-dimensional item parameters such as discrimination and difficulty. As mentioned before,  we include Item Response Theory (IRT) \cite{Johns2006,Brzezinska2020} and Multidimensional IRT (MIRT) \cite{Reckase2009} into the review even if they were proposed before the term \textit{cognitive diagnosis} appeared.

\subsubsection{Item Response Theory (IRT)} \label{sec:irt}
\par
\begin{figure}
    \centering
    \includegraphics[width=.9\textwidth]{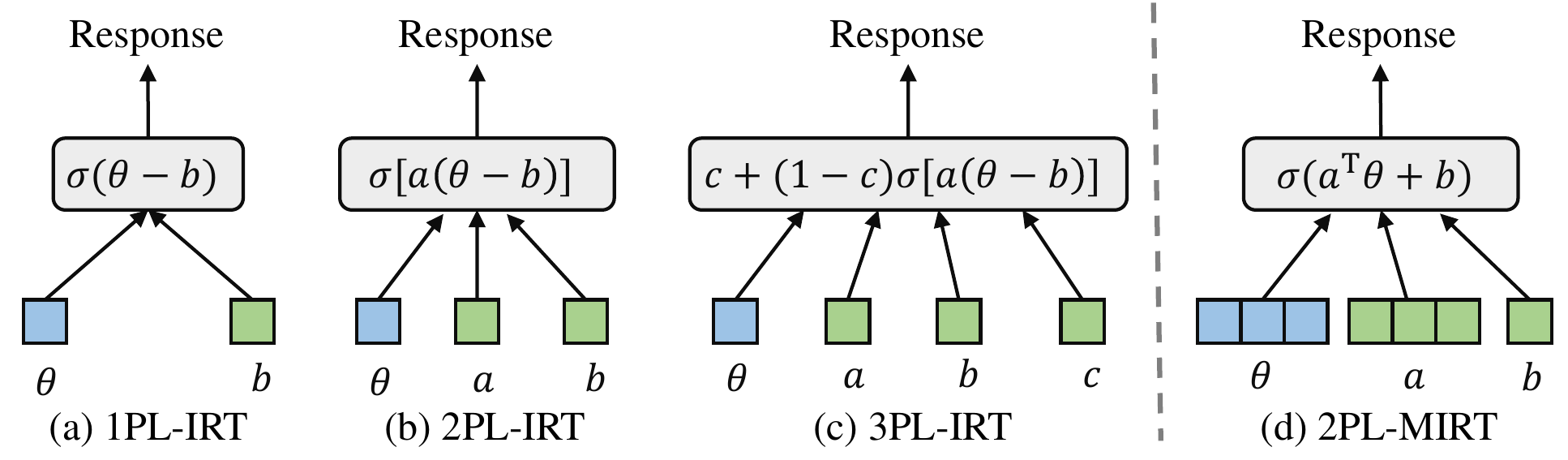}
    \caption{The comparison of representative IRT and MIRT models.}
    \label{fig:irt_mirt}
\end{figure}

 IRT \cite{Johns2006,Brzezinska2020} is one of the most classical latent trait methods for measuring human cognitive status. The core assumption of IRT is that the relation between examinees' responses and their ability levels can be modeled by a continuous mathematical function. i.e., $Pr(r_{ij}=1) = f( \theta_i, \beta_j)$, where $\theta_i$ is a scalar parameter indicating the ability level of examinee $i$, and $\beta_j$ denotes the latent traits of test item $j$. Many IRT-based models are simply called IRT, among which the most representative models include one-parameter logistic IRT (1PL-IRT), two-parameter logistic IRT (2PL-IRT) and three-parameter logistic IRT (3PL-IRT). 1PL-IRT only uses a scalar parameter $b_j$ to capture the difficulty of item $j$. 2PL-IRT adds an extra scalar parameter $a_j$ to indicate the discrimination of item $j$. While 3PL-IRT adds a parameter $c_j$ for item $j$, which is mostly interpreted as the probability of correctly guessing the answer. These models take dichotomous response scores into consideration, i.e., $r_{ij} = 0,1$ indicating incorrect and correct responses respectively. Their formulas are as follows, and Fig.~\ref{fig:irt_mirt} depicts their model structures as generative probabilistic graphical models.
\begin{equation}
    \text{1PL-IRT: } Pr(r_{ij}=1 | \theta_i, b_j) = \sigma(\theta_i - b_j) = \frac{1}{1+e^{-(\theta_i-b_j)}},
\end{equation}
\begin{align}
    \text{1PL-2RT: } & Pr(r_{ij}=1 | \theta_i, a_j, b_j) = \frac{1}{1+e^{-a_j(\theta_i-b_j)}},                   \\
    \text{1PL-3RT: } & Pr(r_{ij}=1 | \theta_i, a_j, b_j, c_j) = c_j + (1-c_j)\frac{1}{1+e^{-a_j(\theta_i-b_j)}}.
\end{align}

\par Despite IRT models represent the ability levels in the simplest scalar form, their excellent mathematical properties (e.g., the convex property of the interaction function) allow them to be applied to a variety of downstream tasks such as computerized adaptive testing. Although IRT models for dichotomous items attract more attention, there have been extended models for polytomous items. For polytomous items, scores can be multiple-graded. For example, for an item whose full score is 10, the obtained scores might be 0, 5, 8 and 10. Researchers have developed various polytomous IRT models such as Partial Credit Rasch Model \cite{Masters1982pcm}, Rating Scale Model \cite{Andersen1997ratingsm} and Graded Response Model \cite{Samejima1969grm}.
 
\subsubsection{Multidimensional Item Response Theory (MIRT)}
MIRT \cite{Reckase2009} extends the ability modeled in IRT to multidimensional cases. Similar to exploratory factor analysis (EFA),  MIRT allows for the exploration of multidimensionality and complex relationships between observed variables and latent traits, particularly in the context of assessments or tests. The simplest form of MIRT is a direct extension from 2PL-IRT and is given as $Pr(r_{ij}=1|\theta_i, a_j, b_j) = 1 / (1+e^{-(a_j^\top\theta_i+b_j)})$,
where $a_j$ denotes item discrimination on multiple dimensions, and $b_j$ is related to item difficulty. In practice, a small dimension is usually enough for MIRT, and each dimension of $\theta_i$ represents a specific ability required to successfully answer the item. However, when using low-dimensional parameters, the ability is hard to explain explicitly. just like factor analysis. Besides relating to EFA and extending the dimension of IRT, MIRT also connects the cognition level paradigm. In the cognition level paradigm, researchers focus more on the fine-grained cognitive states of examinees, such as proficiency levels on pre-defined knowledge concepts. Along this line, da Silva et al. \cite{silva2019qmatrixmirt} introduce the Q-matrix into the interaction function of MIRT to obtain examinees' knowledge-concept-wise latent traits. To ensure the identifiability of MIRT, item discrimination vector $a_j$ of different items are usually rotated to the same value to acquire identifiable estimations of examinees' abilities \cite{Béguin2001mirtfit}.

\subsection{Cognition Level Paradigm}  \label{sec:cognition_level}
\par In the cognitive level paradigm, researchers focus on the estimation of the fine-grained cognitive states of students. For instance, in K-12 course learning, test designers require diagnosing students' proficiency level on knowledge concepts (e.g., the concept \textit{linear function} in mathematics) from their test performances. Therefore, CDMs in the cognitive level paradigm are utilized to estimate student knowledge proficiencies in this scenario. Since such a diagnosis process can be viewed as classifying students to an ``ideal'' proficiency pattern that is most suitable for his/her test performance, traditional cognitive level paradigm-based CDMs are also named as \textbf{Diagnostic Classification Model} (DCM).

\subsubsection{Rule Space Method (RSM) And Its Variations}
\par The RSM proposed by Tatsuoka in the 1980s~\cite{tatsuoka1983rule}, is a statistical modeling approach used for cognitive diagnosis. Compared with IRT, RSM focuses on representing the cognitive processes that individuals use to respond to test items, which is more fine-grained. It seeks to identify the specific cognitive rules or strategies that individuals employ when answering test items. Four steps are included in RSM, i.e., item decomposition, rule specification, rule space construction, and rule space matching.

\par The RSM is a fundamental yet significant method in traditional CD, and is the basis of many DCMs \cite{leighton2004attribute,leighton2007cognitive}. One shortcoming of the traditional RSM is that it views knowledge concepts as independent entities and ignores their hierarchical relationship in the cognitive process (e.g., knowledge dependency). Therefore, Leighton et al. \cite{leighton2004attribute} proposed the Attribute Hierarchy Method (AHM) to address this issue. Compared to the original RSM, the AHM assumes that attributes (i.e., knowledge concepts or skills that are required for students to solve a test item) are organized in a hierarchical structure, which can be represented by an adjacent matrix of attributes. Then the AHM limits the rule space defined in the RSM, such that the mastery of any child attribute should be no less than the mastery of its parent attribute. Then the AHM matches each student into the most similar ideal response patterns, with the corresponding rule as the diagnostic result of the student.

\begin{figure}
    \centering
    \includegraphics[width=.95\textwidth]{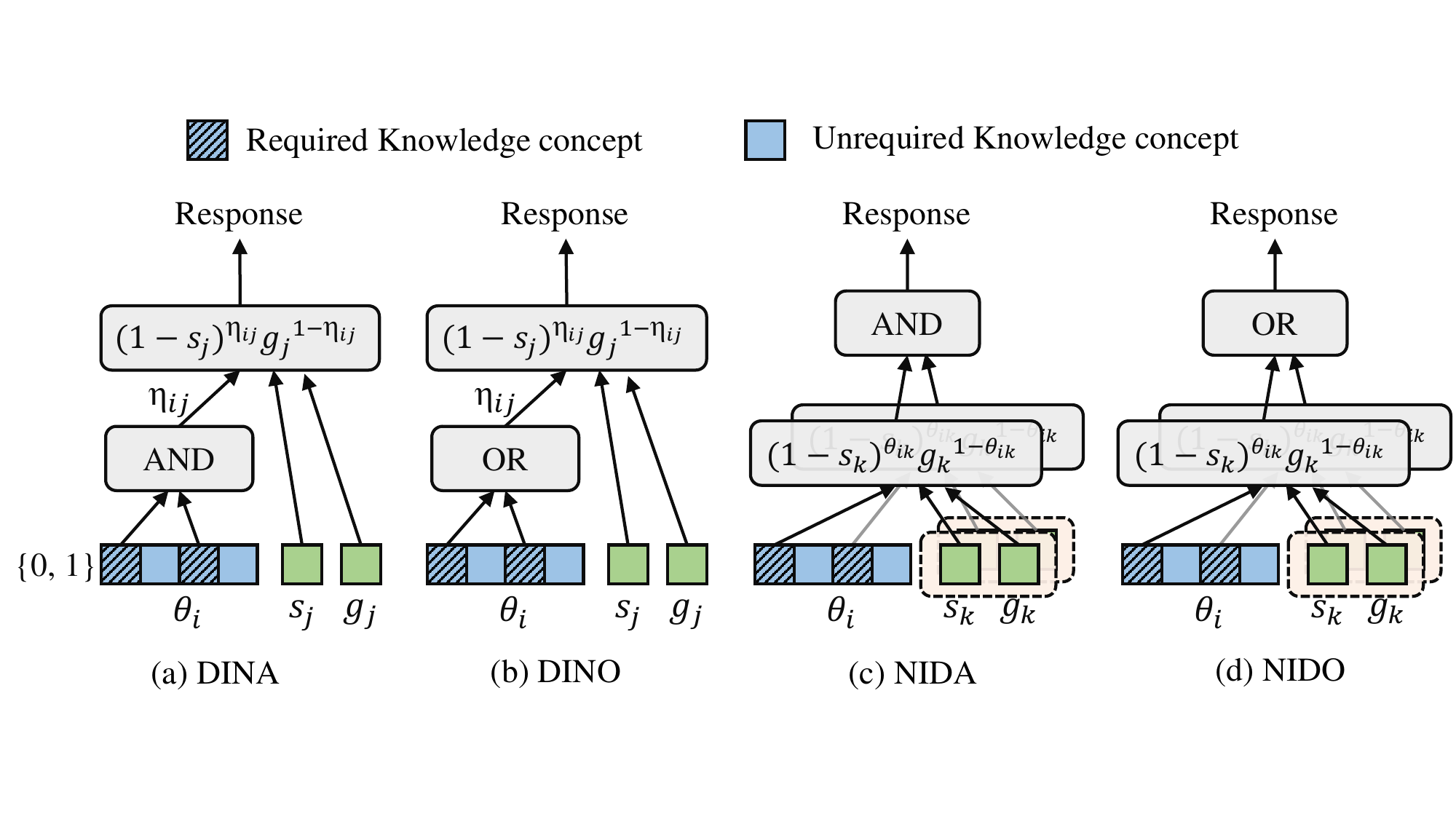}
    \caption{The comparison among DINA, DINO, NIDA and NIDO.}
    \label{fig:DINA_etc}
\end{figure}
\subsubsection{DINA And Relevant CDMs}
\par Deterministic Input, Noisy ``And'' Gate (DINA) model \cite{de2009dina} is a representative and recognized CDM. DINA and its relevant CDMs diagnose students' knowledge concept-wise abilities from binary response data and expert-labeled question-knowledge relationship. The core assumption of DINA is that the proficiency of different knowledge concepts is \textbf{non-compensatory}. That is, the model assumes that mastery of all required attributes is necessary for a correct response, but also allows for the possibility of guessing. In DINA, each student $i$'s ability status is modeled as a binary knowledge mastery pattern vector $\theta_i = (\theta_{i1},\ldots,\theta_{iK})$. Here $K$ denotes the number of knowledge concepts, and the binary value $\theta_{ik} \in \{1, 0\}$ denotes whether or not student $i$ has mastered the knowledge concept $k$. Items are modelled by ``slip'' and ``guess'' parameters and a pre-given binary Q-matrix $Q = (q_1, \ldots, q_J)^\top = (q_{jk})^{J\times K}$. The interaction function of DINA is defined as $Pr(r_{ij}=1|\theta_i,q_j, s_j, g_j) = (1-s_j)^{\eta_{ij}}g_j^{1-\eta_{ij}}$, where $\eta_{ij} = \prod_{k=1}^{K}\theta_{ik}^{q_{jk}}$ is the indicator of whether the student has mastered all required knowledge concepts of the item. The $s_j$ and $g_j$ denote respectively the ``slip'' and ``guess'' parameters of item $j$. Then the goal of the DINA model is to estimate the psychological factors including student cognitive state $\theta_i$, item parameters $s_j$ and $g_j$ given observed binary response scores.

There are some representative CDMs similar to DINA but different in the assumption. Fig. \ref{fig:DINA_etc} demonstrates a comparison among DINA and the following examples. The Deterministic Input, Noisy ``Or'' Gate (DINO) model \cite{templin2006dino} assumes the proficiency of different knowledge concepts to be \textbf{compensatory}. That is to say, the examinee is supposed to correctly answer the item if at least one required knowledge concept is mastered and there is no slipping. Compared to DINA, DINO has a more relaxed assumption and is suitable for certain circumstances such as items with multiple solving strategies. One risk of DINO is an over-estimation of students' ability levels. The Noisy Inputs, Deterministic "And" Gate (NIDA) model \cite{maris1999nida} and Noisy Inputs, Deterministic "Or" Gate (NIDO) model assume that examinees may slip on their mastered knowledge concepts or guess on unmastered knowledge concepts. These noisy mastery patterns then compose the response in compensatory and non-compensatory ways respectively.

\subsubsection{General Diagnostic Model (GDM)}
\par The General Diagnostic Model (GDM) \cite{vondavier2005gdm} is a general framework that subsumes many classical and well-known CDMs like DINA, IRT and MIRT \cite{vondavier2014gdm, bradshaw2014gdm}. As a general framework, GDM is suitable for various real-world scenarios, including but not limited to dichotomous/polytomous response scores, binary/continuous/polytomous ordinal knowledge mastery levels, etc. We introduce the general form of GDM in this section.
\par Formally, let $\theta$ be a K-dimensional skill profile consisting of polytomous or dichotomous skill attributes $\theta_k (k = 1,\ldots, K)$. Then the probability of a polytomous response score $x\in\{0,\ldots,m_j\}$ to item $j$ under the GDM with an individual with skill profiles $\theta$ is defined as
\begin{equation}
    Pr(r_j = x|\theta = (\theta_1,\ldots,\theta_K)) = \frac{\exp\left[\beta_{jx} + \sum_{k=1}^{K}\gamma_{jxk}h_j(q_{jk},\theta_k)\right]}{1+\sum_{y=1}^{m_j}\exp\left[\beta_{jy} + \sum_{k=1}^{K}\gamma_{jyk}h_j(q_{jk},\theta_k)\right]},
\end{equation}
where $\beta_{jx}$ and $\gamma_{jxk} (j = 1,2,\ldots,J)$ are estimable item parameters. Each element $q_{jk}, j = 1,2,\ldots J, k = 1,2,\ldots, K$ of the Q-matrix is a constant, as in other DCMs like DINA. The helper function $h(\cdot,\cdot)$ maps $q_{jk}$ and $\theta_k$ to a real number, which considers the fact that the knowledge profile $\theta$ might be polytomous. By elaborately designing the form of the parameter $\gamma$ and the helper function, GDM can be flexibly applied to either binary or polytomous response data. As a constraint latent class model, the parameter estimation of GDM is usually done with expectation-maximization (EM) algorithm \cite{dempster1977em,huo2014em}.

Due to its generality and flexibility, there are also many other versions for GDM to adjust some special scenarios, like mixture distribution extensions of GDM which consider the ability prior of student groups, and hierarchical extensions of GDM which consider the multilevel distribution of student abilities. Indeed, many traditional CDMs, including CDMs in the ability level paradigm like IRT and MIRT \cite{bradshaw2014gdm}, and CDMs in the cognitive level paradigm like LCDM and DINA \cite{vondavier2014gdm}, have been proven to be special cases of GDM. 

\subsubsection{Other Traditional CDMs}
\par Besides representative CDMs introduced above, there are also some other traditional CDMs that focus on different research challenges in the context of educational measurement. For instance, the Reparameterized Unified Model (RUM) \cite{hartz2002rum}, as a refinement of DINA, aims to construct a cognitive diagnosis assessment system that includes DCM models, estimation procedure, classification algorithm and model-and-data checking function. De La Torre \cite{de2011generalized} proposed the G-DINA, as a general cognitive diagnosis framework similar to GDM, which subsumes many existing CDMs like DINA. Henson et al. \cite{henson2009lcdm} proposed the Log-Linear Cognitive Diagnosis Model (LCDM) based on GDM, which integrates the log-linear model into the calculation of $h(\cdot,\cdot)$. Therefore, LCDM is a special case of GDM and focuses more on the interaction between students and items \cite{vondavier2014lcdm}. Embretson and Yang \cite{embretson2013mltmd} proposed the Multicomponent Latent Trait Model for Diagnosis (MLTM-D), which aims to diagnose hierarchically structured skills or knowledge concepts. In a word, traditional CDMs are usually the combinations of psychometric assumptions and statistical methods.
\section{Machine learning-based Cognitive Diagnosis Models} \label{sec:ML_CD}
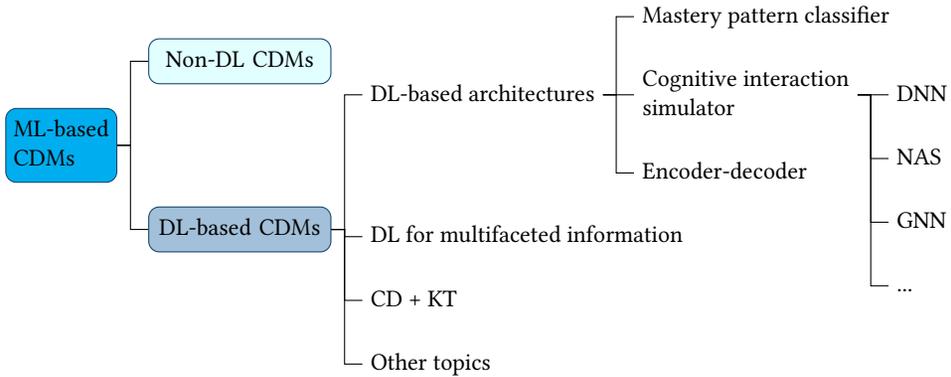
\begin{figure}
    \centering
    \begin{forest}
    for tree={
        scale=0.9,
        grow'=east,
        align=left,
        text centered,
        rectangle,
        rounded corners,
        parent anchor=east,
        anchor=west,   
        child anchor=west,
        edge path={  
            \noexpand\path [draw, \forestoption{edge}]  
            (!u.parent anchor) -- +(5pt,0) |- (.child anchor)\forestoption{edge label};  
        },
    },
    where level=1{text width=7em}{},
    [ML-based\\ CDMs, draw=hidden-draw, fill=cyan 
        [Non-DL CDMs, draw=hidden-draw, fill=LightCyan]
        [,phantom]  
        [,phantom]
        [,phantom]
        [DL-based CDMs, draw=hidden-draw, fill=SteelBlue!50
            [DL-based architectures
                [Mastery pattern classifier]
                [Cognitive interaction\\ simulator, calign=first
                    [DNN]
                    [NAS]
                    [GNN]
                    [...]
                ]
                [Encoder-decoder, align=center]
            ]
            [DL for multifaceted information]
            [CD + KT]
            [Other topics]
        ]
    ]
    \end{forest}
    \caption{A taxonomy of machine-learning based cognitive diagnosis models.}
    \label{fig:nn_cd_taxonomy}
\end{figure}

In recent years, with the development of AI-based education, cognitive diagnosis has raised the attention of researchers from computer science, especially artificial intelligence. CDMs based on machine learning (ML), especially deep learning (DL), are consequently proposed \cite{liu2021towards}. With the advantages of better fitting ability and more flexible structures to make use of different types of educational data (e.g., response, item content, knowledge concept structure), machine learning-based CDMs have achieved much success, leading to a new trend of research. We roughly classify these works according to their proposed time and technology/target in Fig. \ref{fig:nn_cd_taxonomy}.

\subsection{Non-deep-learning Models} \label{sec:ml_nondeep}
In the earlier works, there were several attempts using clustering algorithms to classify examinees into different clusters, where each of the clusters represents a type of knowledge mastery pattern. For example, Chiu~\cite{chiu2009cluster} adopted K-means clustering combined with hierarchical agglomerative cluster analysis, Guo et al.~\cite{guo2020spectral} adopted spectral clustering algorithms. Support vector machine (SVM) was also used in several studies for cognitive diagnosis~\cite{zheng2010application,liu2018application}, where examinees' response logs are input as features and possible knowledge mastery statuses are predicted by SVM. Supervised training data including the known knowledge status is required for SVM-based methods, which limits the practicality of these models. Collaborative filtering methods such as matrix factorization were also adopted to solve the cognitive diagnosis problem in education \cite{toscher2010collaborative,thai2010recommender,thai2015multi}. However, as these models focus on predicting students' performance instead of diagnosing students' knowledge proficiencies, the estimated student parameters are not explicitly explainable. Wu and Liu et al. \cite{wu2015cognitive,liu2018fuzzy} proposed FuzzyCDF which integrates the fuzzy set to handle both objective and subjective test items. Wu et al. \cite{wu2020variational} introduced a variational Bayesian inference algorithm for IRT, which provides a faster and more accurate human ability estimation compared to traditional IRT, especially on large-scale datasets.

\subsection{Deep learning-Based Models} \label{sec:ml_deep}
Integrating deep learning methods has been a new trend in cognitive diagnosis. According to the model architectures and their emergence time, deep learning-based CDMs can be generally classified into \textbf{mastery pattern classifier}, \textbf{cognitive interaction simulator}, and \textbf{encoder-decoder-based} architectures. In \S\ref{sec:dl_architecture} we will review these types of CDMs in detail, following an exploration of multifaceted information (\S\ref{sec:multifaceted}) and other topics in cognitive diagnosis research (\S\ref{sec:dl_others}). 

\subsubsection{DL-based Architectures} \label{sec:dl_architecture}

\begin{figure}
    \centering
    \includegraphics[width=.95\textwidth]{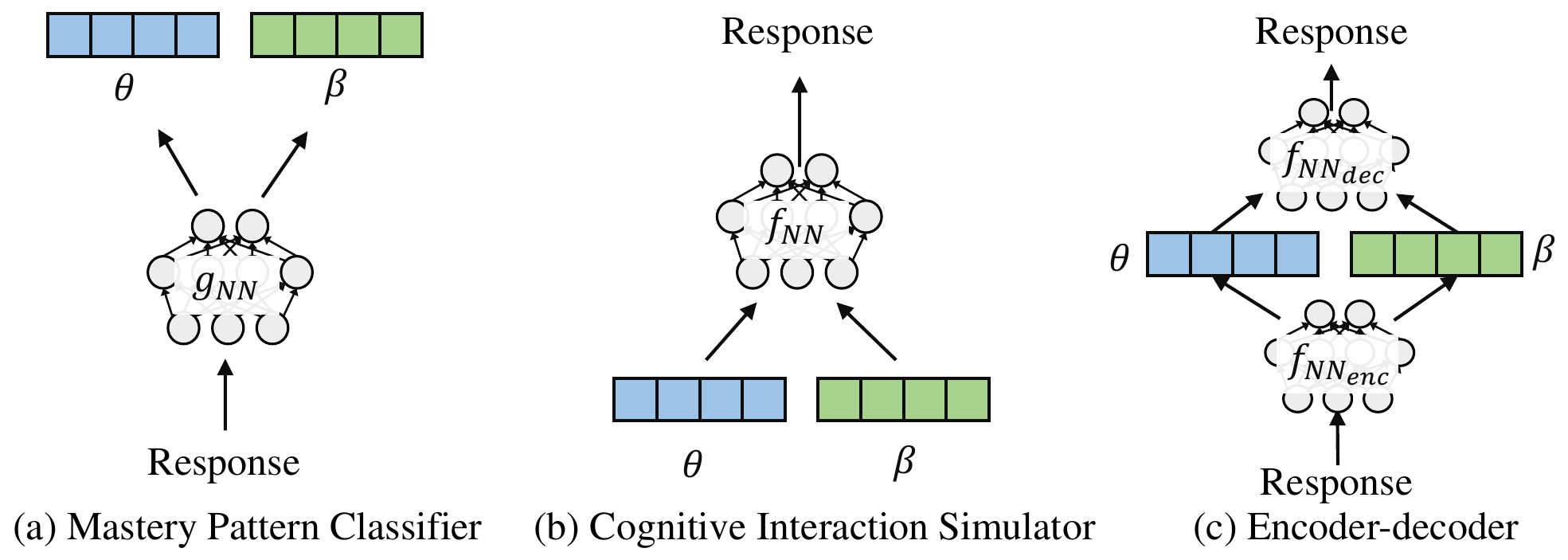}
    \caption{The comparison of deep learning-based cognitive diagnosis models.}
    \label{fig:dl_framework}
\end{figure}

\paragraph{\textbf{(1) Mastery Pattern Classifier}}~{}\newline 
\indent Artificial neural networks were initially adopted in cognitive diagnosis in a reverse manner compared to traditional models, as depicted by Fig. \ref{fig:dl_framework} (a). Typically, as introduced in the Introduction, CDMs simulate the cognitive mechanism of the human's item-answering process. Therefore, the goal of cognitive diagnosis, i.e., humans' ability levels, is actually at the input side of traditional models because it's the cause of human responses. The ability evaluation is actually done through model training instead of model inference. In contrast, Gierl et al. \cite{gierl2008using} proposed a reversed model structure based on neural networks, which takes the examinee's response pattern (i.e., a binary vector that indicates his/her responses) as input, and directly outputs the examinee's attribute pattern (i.e., a vector that indicates his/her mastery on each knowledge concept). As a result, the CDM becomes a classification model, which can be abstracted as:
\begin{equation}
	\theta = g_{NN}(Response, \Omega).
\end{equation}
As empirical response data with the examinee's true ability levels is unavailable, the model is trained with simulated data, which is generated using traditional CDMs such as AHM. Similarly, Cui et al. \cite{cui2016statistical} adopted the self-organizing map to construct the classification model for ability evaluation. The main advantage of these methods is that the ability evaluation can be done with model inference after model training, even if the examinee is out of the training data. However, as pointed out by Briggs et al \cite{briggs2017challenges}, the performance of these models is limited by the CDMs that generate the training data. When the items and/or data generation model is flawed, the trained CDM will naturally incorporate those flaws.  Moreover, the ability estimation can be unstable after multiple times of model training.

\paragraph{\textbf{(2) Cognitive Interaction Simulator}}~{}\newline 
\indent Most deep learning-based CDMs still follow the traditional practice, i.e., model the cognitive interaction during the item answering process like a simulator, as depicted by Fig. \ref{fig:dl_framework} (b). The differences mainly lie in how the interactions are modeled. As pointed out by Wang et al.~\cite{wang2020neural}, traditional psychometric-based CDMs rely on domain experts to design interaction functions for predicting examinees’ responses. Although this approach offers high interpretability, it is costly, and the fixed
function form often leads to weak fitting and generalization capabilities. Wang et al.~\cite{wang2020neural} made an initial break through along this line. They abstracted the cognitive factors involved in the process of answering questions and attributed them to student factors, exercise factors, and interaction functions. A neural network-based framework called NeuralCD was proposed, which can be generally formulated as follows:
\begin{equation}
	Response = f_{NN}(\theta, \beta, \Omega). \label{eq:complex_modeling_ob}
\end{equation}
Here, $\theta$ represents the examinee's ability level and is modeled with a multi-dimensional continuous vector, where the value of each entry indicates the proficiency of a specific knowledge concept. The item parameters $\beta$ here can be the knowledge difficulty, item discrimination, etc. The main difference between NeuralCD and psychometric-based CDMs is that NeuralCD introduced the \textbf{data-driven} strategy to learn the interaction function with neural networks from actual response data. The \textbf{monotonicity} assumption was adopted together with the knowledge relevancy vector to ensure the explainability of diagnostic results, i.e., the estimated $\theta$. Wang et al. also illustrated the generality and extensibility of NeuralCD. The simplicity, robust generalizability, and psychological interpretability based on the monotonicity assumption make NeuralCD highly attractive. Consequently, NeuralCD has inspired the emergence of quite a few DL-based CDMs.

\textbf{Deep neural network (DNN).} The DNN is the most straightforward deep learning method to construct the interaction functions. Wang et al.~\cite{wang2020neural} have provided a basic implementation based on NeuralCD called NeuralCDM. 
Formally, NeuralCDM reconstructs the response of each learner $i$ to the test item $j$ with the formula $Pr({r}_{ij}=1)=f_{DNN}(q_{j}\circ(\theta_{i} - b_j)*a_j)$, where the interaction function $f_{DNN}$ consists of multiple fully connected layers. The ability status of learner $i$ is represented with a $K$-dimensional vector, $\theta_{i} \in \mathbb{R}^{K}$, where $K$ denotes the total number of knowledge concepts corresponding to all test items. Each dimension $\theta_{ik}$, ranging from 0 to 1, represents the proficiency level of learner $i$ on concept $k$. In NeuralCDM, the test item is considered as a difficulty parameter $b_{j} \in \mathbb{R}^{K}$ and a discrimination parameter $a_{j} \in \mathbb{R}^{1}$, similar to MIRT. 
$q_{j} \in \mathbb{R}^{K}$ is the $j$th row of the Q-matrix indicating the knowledge concepts required by item $j$. The symbol $\circ$ denotes the element-wise product. Notably, NeuralCDM strictly adheres to the monotonicity assumption by restricting the layer weights to be nonnegative.

Some deep learning-based CDMs are direct extensions of NeuralCDM. Wang et al.~\cite{wang2022neuralcd} proposed KaNCD to model the latent knowledge associations with the aim of mitigating the low knowledge coverage problem as well as improving the diagnostic performance. Ma et al.~\cite{ma2022knowledge} proposed KSCD that introduces additional parameters for knowledge concepts to better capture the relationship among students, items, and knowledge concepts. Meanwhile, Li et al.~\cite{li2022neuralncd} added the guessing and slipping parameters of test items and replaced the first-layer of the NeuralCDM interaction module with IRT to enhance the interpretability while maintaining data-driven generalization, effectiveness, and efficiency. Cheng et al.~\cite{cheng2021neural} added a knowledge point importance vector to the input layer of NeuralCDM. Similarly, Li et al.~\cite{li2022cognitive} added parameters to depict the impacts of different knowledge concepts, as well as the guessing and slipping factors, thereby proposing the CDMFKC model. Wang et al.~\cite{wang2021using} improved NeuralCDM by aggregating the knowledge concepts by converting them into a graph structure and only considering the leaf node of the knowledge concept tree. 
Some extensions made use of extra information, such as the text content~\cite{wang2022neuralcd}. We will introduce them in Section~\ref{sec:multifaceted}.

\textbf{Neural architecture search (NAS).}
In addition to widely used neural networks, some studies have opted for the more complex neural architecture search (NAS)-based approach to construct the interaction function $f_{NN}$, offering an intriguing alternative. Despite this shift, these endeavors still operate within the NeuralCD framework and can thus be represented using Eq.~(\ref{eq:complex_modeling_ob}). The primary distinction lies in implementing the interaction function $f_{NN}$ using NAS instead of conventional neural networks.
For example, Yang et al.~\cite{yang2023evolutionary,yang2023designing} leverage evolutionary NAS to implement the interaction function. They define the cognitive diagnosis task as a search space within NAS and design various algorithms to search for the optimal solution, aiming to automatically fit the interaction between learners and items.

\textbf{Graph neural network (GNN).} Some researchers have realized that the correlation between learners, items, and even knowledge concepts can form a graph structure, and therefore incorporated GNN into their cognitive diagnostic models. In these works, GNN was mainly used to either improve the embeddings of learners, items, and knowledge concepts~\cite{liu2021graph,meng2022dual} or model the propagation of the influence among the mastery of different knowledge concepts~\cite{gao2021rcd,su2022graph}. Meanwhile, the interaction among the learners, items, and knowledge concepts was still modeled with DNN. We will review more details in Section \ref{sec:multifaceted}.

\textbf{Others.} There are a few studies trying to combine some traditional psychometrics-based CDMs with deep learning. For example, Gao et al.~\cite{gao2022deep} combines the IRT, DINA and neural networks, and predicts learners' scores on both objective and subjective items. Wang et al.~\cite{wang2023unified} took IRT, DINA, HO-DINA, and MIRT into consideration and fused them in two ways with neural networks.

\paragraph{\textbf{(3) Encoder-Decoder-based Architecture}}~{}\newline 
\indent A few recent studies have raised attention to encoder-decoder-based CDMs. If we regard the first type of architecture (mastery pattern classifier) as the encoder and the second type (cognitive interaction simulator) as the decoder, then the new architecture can be seen as the combination of the above (Fig. \ref{fig:dl_framework} (c)). Encoder-decoder-based CDMs take responses (and maybe more types of data in the future) as input, encode them into examinees' explainable ability vectors and item parameters, and then decode them to reconstruct the input responses. It can be formalized as follows:
\begin{equation}
	Response = f_{NN_{dec}}(\theta, \beta, \Omega_{dec}|\theta, \beta \leftarrow f_{NN_{enc}}(Response, \Omega_{enc})), \label{eq:enc_dec_ob}
\end{equation}

Such architecture overcomes the shortcomings of the mastery pattern classifier architecture as it can be trained directly using real-world datasets. Moreover, after model training, the diagnosis process can be conducted \textbf{inductively} which means that examinees who do not appear in the training data can be directly diagnosed based on their response data using the encoder. Along this line, Li et al.~\cite{li2024towards} proposed a response-proficiency-response paradigm called ID-CDF, where the encoder module is implemented with a simple yet effective DNN. Similarly, Liu et al.~\cite{liu2024inductive} also proposed an inductive encoder-decode-like CDM called ICDM, where the authors constructed a student-centered graph based on the response data and Q-matrix, and encoded the nodes (i.e., students, questions, and knowledge concepts) into embedding vectors that can be transformed to explainable student's ability levels. In addition, in ~\cite{li2024towards}, the authors raised concerns about the inherent non-identifiability and explainability overfitting issues of the traditional decoder-like architecture. These issues can negatively impact the quantification of learners' cognitive states and the quality of web learning services. The superiority of ID-CDF in solving the above problems has been verified in the paper. Chen et al. proposed DCD~\cite{chen2024disentangling}, which maps cognitive parameters and test item traits into distributional forms, utilizing a variational autoencoder as the interaction function to enhance the model's generalization capability. The ICD model proposed by Qi et al.~\cite{qi2023icd} basically follows the encoder-decoder architecture. A data enhancement approach was proposed so that each learner's responses were divided differently multiple times.

\subsubsection{Integration of Multifaceted Information.} \label{sec:multifaceted}
Despite significant progress in cognitive diagnosis interaction function technology, the bottleneck of cognitive diagnosis arises from the initialization of diagnosis factors (learner features and test item features) solely based on their IDs. Thus, researchers have begun exploring ways to enhance the expressive ability of diagnosis factors with multifaceted information, including side-information and domain priors, aiming to further improve the interpretability and performance of diagnosis models.

Let $\mathbf{X}$ denote the multifaceted information. We hence model the CDM that introduces multifaceted information as follows:
\begin{equation}
	Response = f_{NN}(\theta, \beta, \Omega|\theta \leftarrow f_{NN_{user}}(\mathbf{X}, \Omega_{user}), \beta \leftarrow f_{NN_{item}}(\mathbf{X}, \Omega_{item})), \label{eq:side_information_ob}
\end{equation}
where $f_{NN_{user}}$ and $f_{NN_{item}}$ are feature extraction functions to extract useful clues from multifaceted information to generate solid diagnosis factors, i.e., learner cognitive traits and test item traits. $f_{NN}$ is the interaction function for the diagnosis prediction.
The multifaceted information $\mathbf{X}$ commonly utilized in cognitive diagnosis can be categorized into three types as follows.

\textbf{Learner-side information.} From the perspective of learners, extra information typically includes learner features or profiles, such as age, gender, behaviors, and preferences, as well as contextual information like school details, family income, and parental occupation, all of which are relevant to the learner. The learner-side information can provide richer insights than mere IDs. Zhou et al.~\cite{zhou2021modeling} leveraged learner features (e.g., gender, age, region) as the initial profile and contextual information related to the learners (e.g., school details and parents' occupation) to uncover implicit relations between learners' contextual information and their performance in practice. This study not only enhances diagnostic performance but also sets the groundwork for subsequent research on fairness in education~\cite{zhang2024understanding}.

\begin{figure}
    \centering
    \includegraphics[width=.95\textwidth]{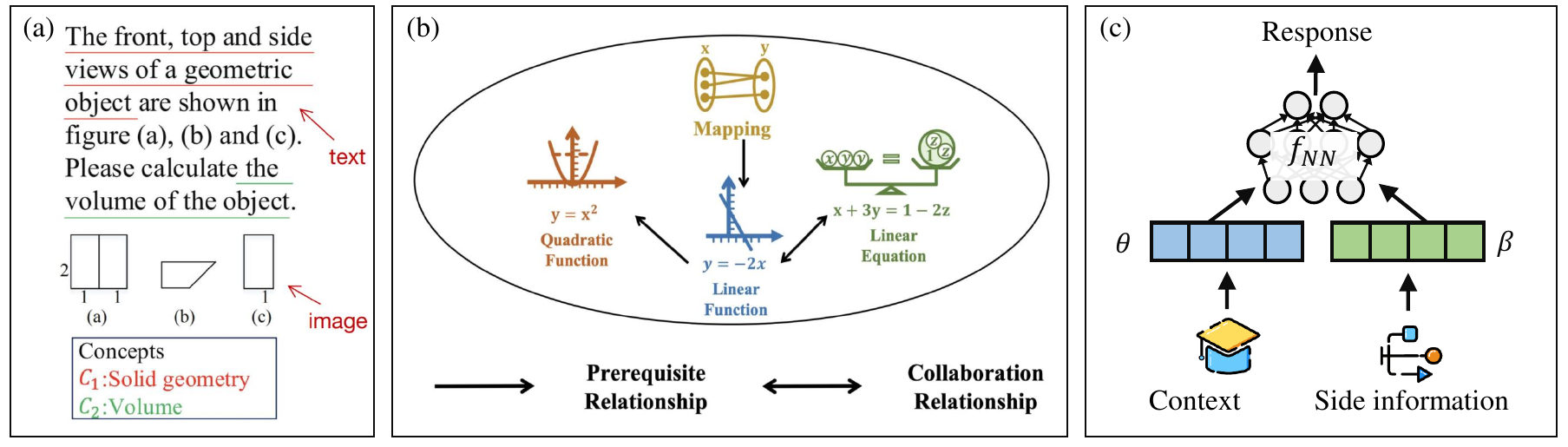}
    \caption{(a) Examples of item-side information; (b) An example of knowledge concept graph with prerequisite and collaboration relations; (c) DL-based CDMs integrating multifacet information.}
    \label{fig:features}
\end{figure}

\textbf{Item-side information.} In addition to relevant knowledge concepts/skills, commonly employed item-side information encompasses the item contents, such as texts and images (Fig.~\ref{fig:features} (a)), as well as exceptional factors like guessing and slipping, which offer detailed semantic cues to represent item traits. Song et al.~\cite{song2023deep}, Cheng et al.~\cite{cheng2019dirt}, Gao et al.~\cite{gao2024zero}, and Wang et al.~\cite{wang2022neuralcd} extracted item difficulty and discrimination from text or image content, enhancing the extensibility of CDMs to cold-start items. Gao et al.~\cite{gao2022deep} established semantic relationships between item text and knowledge concepts, enhancing the interpretability of cognitive diagnosis.

\textbf{Relational graph-based information.} Many studies introduce relational graphs based on the educational priors, such as knowledge concept (KC) graphs (as depicted in Fig.~\ref{fig:features} (b)) and item-concept association graphs, to enhance the representations of both learners and items. Gao et al.~\cite{gao2021rcd} and Su et al.~\cite{su2022graph} modeled the heterogeneous graph structures of learner-item-knowledge to fully explored higher-order interaction relationships between the nodes and the dependency relationships between knowledge concepts within a concept map, thus enhancing the representation of learner cognitive states and item features. Li et al~\cite{li2022hiercdf} proposed the HierCDF framework to model the influence of hierarchical knowledge structures on cognitive diagnosis. Song et al.~\cite{song2023deep} focused on the effective fusion of knowledge concept maps with knowledge concept dependency relations and item features. Jiao et al.~\cite{jiao2023neural} revealed the relationships between knowledge concepts and items, as well as concept dependency relations within the knowledge concept map, enhancing the representation of item and learner features.

\subsubsection{Combination of cognitive diagnosis and knowledge tracing}
As mentioned in \S\ref{sec:preliminary}, cognitive diagnosis assumes constant ability, which is a big difference compared to knowledge tracing. This assumption is reasonable in circumstances when examinees' ability is stable in a relatively short time period. However, there are indeed situations that need to model the changes in examinees' ability levels. For example, at an online learning platform, a learner may receive exercises related to a certain knowledge concept multiple times to practice. This type of learning process involves changes in learners' abilities, which the learners and maybe other platform users care about.

Due to the lack of interpretability of traditional knowledge tracing models, some researchers combined cognitive diagnosis models with knowledge tracing. For example, Zhang et al.~\cite{zhang2021gkt} developed Gated-GNN to trace the student-knowledge response records and to extract students' latent traits, and used IRT to predict the probability of students answering exercises correctly. Gan et al.~\cite{gan2022knowledge} and Li et al.~\cite{li2022dkt} combined the key-value memory network with IRT. Wang et al.~\cite{wang2023dynamic} proposed a Dynamic Cognitive Diagnosis (DynamicCD) approach. This work provided a relatively extensive discussion about what types of educational priors from cognitive diagnosis models can be integrated with knowledge tracing, how they are integrated, and what influences they bring. The proposed approach provides a more nuanced understanding of learners' evolving knowledge states, as well as a more accurate prediction of learners' performance. Ma et al.~\cite{ma2022reconciling} proposed a continuous time based Neural Cognitive Modeling (CT-NCM) that combines the neural Hawkes process with CDMs, thereby integrating the dynamism and continuity of knowledge forgetting into the learning process modeling. Liu et al.~\cite{liu2024automated}  proposed a two-stage method to automatically discover symbolic laws governing skill acquisition from learners' sequential response data. In the first stage of their method, a transformer-like module is used to encode learners' sequential response data and then combined with 3PL-IRT to predict the scores. In the second stage, symbolic regression is used to extract core patterns from the trained deep-learning regressor into algebraic equations, resulting in symbolic rules. It is worth pointing out that, the combination of cognitive diagnosis and knowledge tracing might be seen as an \textbf{extension of the encoder-decoder-based architecture} of CDMs. The difference mainly lies in that the encoder is from the knowledge tracing academia which better models the sequential data and tracks the evolution of knowledge status.

\subsubsection{Other Issues.} \label{sec:dl_others}
Considering the practical demands of real-world scenarios, recent studies have been focusing on developing cognitive diagnosis methods tailored for specific application contexts. These contexts often come with their own complexities and data constraints, necessitating specialized approaches to effectively address the unique challenges they present.

\textbf{Cognitive diagnosis under limited data scenarios.} The practice data of learners in real learning scenarios often face limitations. For instance, the self-driven nature of online learning leads many learners to selectively engage with items they excel in or practice irregularly, resulting in a 'sparse issue' that introduces biases into their diagnosis results. To address this challenge, Yao et al. \cite{yao2023exploiting} analyzed the features of both interacted and non-interacted items and designed an item-aware partial order constraint to guide cognitive diagnosis modeling.

Moreover, in real learning platforms, new courses are frequently introduced where learner data is unavailable, presenting a cold-start problem. Traditional CDMs heavily rely on abundant learner practice data, which becomes scarce in such scenarios. In response, Gao et al. \cite{gao2023leveraging} proposed a unified knowledge concept graph modeling approach to bridge cold-start scenarios with mature scenarios possessing rich data, transferring modeling experience from existing contexts to cold-start situations. Additionally, Gao et al. \cite{gao2024zero} developed a training-fine-tuning approach leveraging pre-training to extract universal cognitive representations from existing scenarios. By fine-tuning a superior cognitive diagnosis model with a small amount of learner practice data in cold-start scenarios, this method effectively overcomes data scarcity challenges.

\textbf{Fairness in cognitive diagnosis.}
Fairness has become a prominent and pressing issue in education, with recent years witnessing a surge in research focused on fair cognitive diagnosis modeling. This is especially significant as learners' diagnostic results produced by CDMs can be impacted by various sensitive attributes, such as region or socio-economic background during the model training. Recognizing this, Zhang et al.~\cite{zhang2024understanding} delved into the fairness in cognitive diagnosis and uncovered instances of unfairness in prior CDMs. An adversarial-based cognitive diagnosis framework was proposed to eliminate sensitive information from user vectors, thereby ensuring fairness in the diagnostic process.
Furthermore, Zhang et al.~\cite{zhang2024path} argued that sensitive attributes of students can also provide valuable information, and only shortcuts directly linked to the sensitive information should be eliminated. To accomplish this objective, they utilized causal reasoning and developed a path-specific causal reasoning framework for fairness-aware cognitive diagnosis.

\textbf{Group-level cognitive diagnosis.} While most CDMs have primarily focused on individual-oriented modeling, real learning scenarios frequently entail group teaching, such as classroom settings. This necessitates the development of group-level cognitive diagnosis methods to enhance teaching efficiency. In response to this need, methods based on psychometrics-based CDMs have been proposed, which either extend IRT to combine with response sampling methods~\cite{reise2006application,birenbaum2004diagnostic} or average the diagnosed individual abilities~\cite{agrawal2014grouping,liu2016collaborative}. Recent studies have been exploring this issue using deep learning methods. Huang et al. \cite{huang2021group} proposed an efficient group-level diagnosis method based on educational priors. By considering the collective knowledge and performance of a group, this approach offers insights that can inform instructional strategies and interventions tailored to group dynamics.
Further advancing group-level diagnosis, Liu et al. \cite{liu2023homogeneous} integrated homogeneous relations among learners to enhance diagnosis performance. By leveraging similarities and shared characteristics among learners within a group, this method offers a more nuanced understanding of group-level cognitive profiles. These advancements hold promise for optimizing teaching practices and facilitating collaborative learning experiences in educational settings.

\textbf{Data privacy in cognitive diagnosis.} Data privacy is of great concern nowadays. Learners' behavioral data on learning platforms could be private data that is not allowed by either learners or platform administrators to share. In response to such data protection policies, Wu et al.~\cite{wu2021hierarchical} and Liu et al.~\cite{liu2023federated} proposed federated learning-based user model approaches and applied them to cognitive diagnosis. These approaches achieved comparable diagnostic performances with local training approaches, with much less risk of data leakage.

\textbf{Cognitive diagnosis in adversarial scenarios.} In some applications of cognitive diagnosis, the data is not limited to examinees' responses to test items, but also the outcome of confrontation between individuals, or even between teams. For example, Gu et al.~\cite{gu2021neuralac} proposed a NeuralAC model to measure the abilities of MOBA (multiplayer online battle arena) game players, where the cooperation and competition factors should be considered. An et al.~\cite{an2021lawyerpan} proposed a proficiency assessment network
for trial lawyers, which the role of a lawyer in court cases, including cooperation with team members and debate with adversarial lawyers, is modeled.

\textbf{Efficiency in cognitive diagnosis.} As Assumption 1 stated, examinees' ability levels are assumed to be constant in CDMs. When used in online learning platforms, learners' parameters should be updated to keep track of the changes in learner ability. However, frequent re-estimation of parameters can be uneconomical. Therefore, Tong et al.~\cite{tong2022incremental} theoretically discussed when a CDM should be updated and how to incrementally update it, thereby proposing an incremental cognitive diagnosis (ICD) method.

Finally, we provide a summary of the descriptive characteristics of main machine learning-based CDMs in Table \ref{tab:ml_CDM_compare}.
\begin{table}[tb]
    \centering
    \caption{A summary of descriptive characteristics of main machine learning-based CDMs. MPC, CIS, ED and KC are the abbreviations of \textit{Mastery Pattern Classifier}, \textit{Cognitive Interaction Simulator}, \textit{Encoder-decoder-based Architecture} and \textit{Knowledge Concept} respectively. The knowledge status provided by several CDMs depends on base CDMs because they are frameworks that can be applied to different existing CDMs.}
    \label{tab:ml_CDM_compare}
    \resizebox{\textwidth}{!}{%
    \begin{tabular}{c|c|c|c|c|c}
        \hline
         CDM & Architecture & ML basis & Knowledge status & Extra Information & Other topics  \\
         \hline
         SVM\cite{zheng2010application,liu2018application} & MPC & SVM & Binary vector & $\backslash$ & $\backslash$  \\
         MF\cite{toscher2010collaborative,thai2010recommender,thai2015multi} & CIS & MF & Hidden state & $\backslash$ & $\backslash$  \\
         FuzzyCDF\cite{liu2018fuzzy} & CIS & Fuzzy Set & Real-valued vector & $\backslash$ & $\backslash$  \\
         \hline
         \hline
         NN\cite{gierl2008using} & MPC & DNN & Real-valued vector & $\backslash$ & $\backslash$ \\
         SOM\cite{cui2016statistical} & MPC & Self-organized map & Binary vector & $\backslash$ & $\backslash$ \\
         \hline
         NeuralCDM\cite{wang2020neural} & CIS & DNN & Real-valued vector & $\backslash$ & $\backslash$  \\
         KaNCD\cite{wang2022neuralcd} & CIS & DNN & Real-valued vector & $\backslash$ & $\backslash$  \\
         KSCD\cite{ma2022knowledge} & CIS & DNN & Real-valued vector & $\backslash$ & $\backslash$   \\
         NeuralNCD\cite{li2022neuralncd} & CIS & DNN & Real-valued vector & $\backslash$ & $\backslash$   \\
         IK-NeuralCD\cite{cheng2021neural} & CIS & DNN & Real-valued vector & $\backslash$ & $\backslash$   \\
         CDMFKC\cite{li2022cognitive} & CIS & DNN & Real-valued vector & $\backslash$ & $\backslash$   \\
         CDGK\cite{wang2021using} & CIS & DNN & Real-valued vector & $\backslash$ & $\backslash$   \\
         \hline
         NAS-GCD\cite{yang2023evolutionary} & CIS & NAS & Real-valued vector & $\backslash$ & $\backslash$   \\
         EMO-NAS-CD\cite{yang2023designing} & CIS & NAS & Real-valued vector & $\backslash$ & $\backslash$   \\
         \hline
         RCD\cite{gao2021rcd} & CIS & GNN, DNN & Real-valued vector & KC structure & $\backslash$   \\
         Graph-EKLN\cite{meng2022dual} & CIS & GNN, DNN & Real-valued vector & KC structure & $\backslash$   \\
         GCDM\cite{su2022graph} & CIS & GNN, DNN & Real-valued vector & $\backslash$ & $\backslash$   \\
         \hline
         deepCDF\cite{gao2022deep} & CIS & Other & Real-valued vector & Text &  Subjective and objective exercises  \\
         LDM-ID/HMI\cite{wang2023unified} & CIS & Other & Real-valued vector & $\backslash$ & $\backslash$   \\
         \hline
         CNCD-Q/CNCD-F\cite{wang2022neuralcd} & CIS & DNN & Real-valued vector & Text & $\backslash$  \\
         ECD\cite{zhou2021modeling}& CIS & DNN & Real-valued vector & Learner context &  $\backslash$  \\
         HierCDF\cite{li2022hiercdf} & CIS & DNN, Bayesian NN & Depends on base CDM & KC structure & $\backslash$ \\
         CMNCD\cite{song2023deep}& CIS & DNN & Real-valued vector & Text, Image, KC structure & $\backslash$   \\
         QI-NeuralCDM\cite{jiao2023neural}& CIS & DNN & Real-valued vector & KC structure & $\backslash$   \\
         \hline
         TechCD\cite{gao2023leveraging}& CIS & GNN, DNN & Real-valued vector & KC structure &  Limited data scenarios, Cold-start  \\
         Zero-1-to-3\cite{gao2024zero}& CIS & DNN & Real-valued vector & Text, Learner relations &  Limited data scenarios, Cold-start  \\
         EIRS\cite{yao2023exploiting}& CIS & DNN & Real-valued vector & Item-aware partial order  &  Limited data scenarios, Data sparsity  \\
         \hline
         FairCD\cite{zhang2024understanding}& CIS & DNN & Real-valued vector & Learner context &  Fairness  \\
         \hline
         MGCD\cite{huang2021group}& CIS & DNN & Real-valued vector & $\backslash$ &  Group-level  \\
         HomoGCD\cite{liu2023homogeneous}& CIS & DNN & Real-valued vector & Learner relations &  Group-level  \\
         \hline
         HPFL\cite{wu2021hierarchical}& CIS & DNN, Federated learning & Real-valued vector & $\backslash$ &  Data privacy  \\
         AHPFL\cite{liu2023federated}& CIS & DNN, Federated learning & Real-valued vector & $\backslash$ &  Data privacy  \\
         \hline
         NeuralAC\cite{gu2021neuralac}& CIS & DNN & Hidden state & $\backslash$ &  Adversarial scenarios  \\
         LawyerPAN\cite{an2021lawyerpan}& CIS & DNN & Real-valued vector & Text &  Adversarial scenarios  \\
         \hline
         ICD\cite{tong2022incremental} & CIS & DNN & Depends on base CDM & $\backslash$ & Efficiency \\
         \hline
         ID-CDF\cite{li2024towards} & ED & Enc-Dec, DNN & Depends on base CDM & $\backslash$ & Identifiability   \\
         ICDM\cite{liu2024inductive} & ED & Enc-Dec, GNN & Real-valued vector & $\backslash$ & $\backslash$   \\
         ICD\cite{qi2023icd} & ED & Enc-Dec, DNN & Real-valued vector & $\backslash$ & $\backslash$   \\
         DASPM\cite{liu2021graph} & ED & Enc-Dec, GNN & Real-valued vector & KC structure & $\backslash$   \\
         DCD\cite{chen2024disentangling} & ED & Enc-Dec, $\beta$-TCVAE & Gaussian distribution & KC structure & Limited Exercise-Knowledge Labels    \\
         \hline
    \end{tabular}
    }
\end{table}
\section{Parameter Estimation} \label{sec:estimation}
Parameter estimation is responsible for the \textbf{psychological factor estimation} within CDMs, especially the parameters representing examinees' ability levels. This section provides several representative parameter estimation methods employed in cognitive diagnosis, including Expectation Maximization (EM), Markov Chain Monte Carlo (MCMC), and Gradient Descent (GD). Since most of the existing cognitive diagnosis models are simulations of examinees' cognitive processes during item answering, estimating parameters is approximately equivalent to diagnosing examinees' ability status. The main purpose of cognitive diagnosis models is to obtain the estimated parameters representing examinees' abilities instead of predicting the examinees' future performance. Regardless of a few exceptions \cite{gierl2008using,cui2016statistical,wu2020variational}, this is a big difference compared to traditional machine learning whose goal is to train models that can be used to predict the labels of unseen samples.


\subsection{Expectation Maximization} \label{sec:EM}
Expectation Maximization (EM)~\cite{dempster1977maximum} is widely adopted for non-deep-learning CDMs, where $\Omega = \emptyset$, such as IRT \cite{woodruff1996estimation}, MIRT \cite{zhang2004comparison}, DINA \cite{de2009dina} and G-DINA \cite{de2011generalized}. Note that in some papers, it is also called the marginalized maximum likelihood estimation (MMLE). The missing data in the EM algorithm refers to the examinees' ability parameters in the CDM. Therefore, a standard EM algorithm for CDMs is as follows:

\textbf{Expectation (E-step)}: In this step, given the observed responses and the current parameter values, the Q-function is calculated: 
\begin{equation}
    Q[(\beta, \pi)|(\beta^{(s)}, \pi^{(s)})] = \sum_{i=1}^I \log [\int_{\theta_i} Pr(R_i|\theta_i, \beta) Pr(\theta_i | R, \beta^{(s)}, \pi^{(s)}) \mathrm{d}\theta_i],
\end{equation}
where $\pi$ represents the prior distribution of examinees' ability parameters and $\beta$ represents all the item parameters; $\beta^{(s)}, \pi^{(s)}$ represent the current values of $\beta$ and $\pi$ estimated in the last iteration. The prior distribution $\pi$ can be either fixed or estimated during the iterations. A convenient strategy is to discretize the prior distribution when $\theta_i$ is a continuous parameter (e.g., in IRT), which significantly simplifies the calculation of the integral \cite{woodruff1996estimation}.
 
\textbf{Maximization (M-step)}: In this step, the algorithm updates the values of $\beta$ and $\pi$ to maximize the Q-function:
\begin{equation}
   (\beta^{(s+1)}, \pi^{(s+1)}) = \underset{\beta, \pi}{\arg\max}\ Q[(\beta, \pi)|(\beta^{(s)}, \pi^{(s)})].
\end{equation}
The EM algorithm iteratively alternates between the E-step and M-step until convergence. 

In the standard practice of EM algorithm for CDMs, the parameter estimation is a two-stage process. The first stage is item calibration, which estimates item parameters while regarding the examinees' abilities as latent variables following particular prior distribution (i.e., $\pi$). This stage is sometimes conducted using the responses provided by a particular group of examinees and is an especially necessary step for computer adaptive testing (CAT) to construct item banks \cite{van2000computerized}. At the second stage, examinees' ability parameters are estimated taking the previously estimated item parameters as known and fixed.
 
\subsection{Markov Chain Monte Carlo}
MCMC can be thought of as a successor to the standard two-stage practice of EM-based methods, which treats the item and examinee parameters at the same time. Compared to EM-based methods, MCMC-based methods are generally easier to implement and remain straightforward as model complexity increases, at the cost of generally slower execution times \cite{patz1999straightforward}.

The Gibbs sampling approach of MCMC can be straightforwardly used for CDMs with multiple parameters. In case when the distribution is difficult to directly draw samples from, one can integrate the Metropolis-Hastings approach that introduces the acceptance rate into the sampling \cite{patz1999straightforward,de2004higher,liu2018fuzzy}. The parameters can be grouped into blocks to update simultaneously so as to increase the efficiency \cite{gelman1995bayesian}. MCMC was extended to address problems such as missing data, multiple item types, rated responses, and response time \cite{patz1999applications, zhan2018cognitive,shan2020cognitive}.

\subsection{Gradient Descent}
Since deep learning-based CDMs have relatively complicated model structures with more parameters, EM-based and MCMC-based methods are not easily extendable and less efficient for these models. Instead, gradient descent (GD), which has been the mainstream estimation algorithm especially used in deep learning, is adopted to train these models. 

Cross entropy added by some regularization terms is most adopted as the objective function \cite{wang2020neural,gao2021rcd,zhou2021modeling,li2022hiercdf,yang2022novel,ma2022knowledge}. Tong et al. and Yao et al. proposed pairwise objective functions to enhance the monotonicity and leverage non-interactive items \cite{tong2021item,yao2023exploiting}. The optimization can be conducted through mini-batches \cite{hinton2012neural}, and optimizers proposed in the deep learning research are compatible with CDM training, such as Adam \cite{kingma2014adam}, Adagrad \cite{duchi2011adaptive}, and RMSProp \cite{tieleman2012rmsprop}. Except for these generally used methods, research about parameter estimation especially for deep learning-based CDMs is still underexplored. The robustness of the existing GD methods for CDMs, such as stability, uncertainty and explainability, might need further analysis.
\section{Model Evaluation} \label{sec:evaluation}

Model Evaluation is an important stage of validating the effectiveness of models and helping with model selection in real-world applications. Basically, a CDM is expected to provide accurate, explainable and even robust diagnostic results to users. Due to the large differences between traditional psychometrics-based CDMs and machine learning-based CDMs, the evaluation methods tend to be different. We make the summary as follows.

\subsection{Evaluation for psychometrics-based CDMs}
The evaluation methods for psychometrics-based CDMs are relatively abundant. Some frequently discussed aspects include goodness-of-fit, reliability, validity, and uncertainty. Goodness-of-fit evaluates whether accurate (and may also be explainable) diagnostic results can be provided by CDMs, while the rest is about the robustness of the diagnostic results. 

\textit{\textbf{Goodness-of-fit.}} Goodness-of-fit provides a general measure of whether a CDM is capable of fitting the response data well. If a CDM cannot even fit most of the data, it is not likely that the CDM can provide accurate diagnostic results. Various metrics for goodness-of-fit have been proposed, which can be generally divided into relative fit evaluation and absolute fit evaluation. The relative fit evaluation compares the fitnesses of different CDMs on a certain dataset. The number of parameters within a CDM may also be taken into consideration to obtain a balance between fitness and model complexity. Such metrics include AIC, BIC~\cite{vrieze2012model}, DIC~\cite{spiegelhalter2002bayesian}, etc. Absolute fit evaluation aims to assess the extent to which a model adequately fits the observed data. Fit indices such as RMSE or chi-square statistic \cite{bolboacua2011pearson} provide quantitative measures of the discrepancy between the model's predicted values and the observed data. Lower values of these indices indicate a better fit between the model and the data. When choosing a CDM to use in applications, a common practice is to use relative fit metrics to select relatively better fitting CDMs, and use absolute fit metrics to determine the best CDM or identify the misspecification of Q-matrix and CDM~\cite{chen2013relative, hu2016evaluation}.

In addition, quite a lot of works used synthesized datasets for model evaluation. As the groundtruths of synthesized datasets are known, a straightforward way to evaluate the accuracy of diagnostic results is to compare the differences between estimated parameters (denoting examinees' ability levels or item characteristics) and the groundtruths~\cite{de2001impact,de2010factors}. When the estimated ability levels are close to the groundtruths, they are intrinsically explainable. However, such a method cannot be applied to real-world datasets.

\textit{\textbf{Uncertainty/Confidence.}}
It is not realistic to expect that CDMs always provide accurate and reliable diagnostic results, in other words, there exists uncertainty within the diagnostic results. For example, a CDM may not be able to make sure of the actual ability of a learner based on the observed data (even if it outputs an estimated ability parameter ``0.7'' for that learner), while it infers that the ability is most likely in the range of (0.6, 0.8). Therefore, understanding the uncertainty or confidence of diagnostic results is valuable for assessing the reliability of diagnostic outcomes, as it unveils potential error margins or ranges in model predictions. By gaining insights into this uncertainty, decision-makers can better comprehend diagnostic results, factor in potential risks and uncertainties, and make more prudent decisions. Various methods have been proposed to estimate the uncertainty of psychometrics-based CDMs. For instance, Fully Bayesian sampling-based methods \cite{patz1999straightforward} and the multiple imputation method \cite{yang2012characterizing} analyze the uncertainty of IRT and MIRT by examining variations in diagnostic results. Frequentist methods \cite{patton2014bootstrap,rights2018addressing} use standard errors to depict uncertainty. Duck-Mayr et al. \cite{duck2020gpirt} introduce a Gaussian process-based approach for nonparametric IRT models.

\subsection{Evaluation for machine learning-based CDMs}
The development of machine learning-based CDMs has a much shorter history compared to Psychometrics-based CDMs, no matter the research about model evaluation. Due to the big difference in model structures and maybe the background of researchers, the evaluation of machine learning-based CDMs is different.

\textit{\textbf{Accuracy-related evaluation.}} To evaluate whether a CDM provides accurate diagnostic results, most studies about machine learning-based CDMs, especially deep learning-based CDMs, adopted the metrics that are usually used to evaluate regression or classification ML models. Research works of machine learning-based CDMs pay more attention to real-world datasets. However, as the groundtruths of examinees' ability levels are not available, the evaluation is usually indirect. Specifically, each examinee's responses are divided into a training set and a testing set. The diagnosis, that is, the parameter estimation, is conducted based on examinees' responses in the training set. Based on the estimated ability parameters, CDMs are required to predict the examinees' responses in the testing set. The motivation is that, if the diagnostic results are accurate, then the prediction of responses based on them should also be accurate. Metrics from both regression tasks and classification tasks have been adopted, such as the root mean square error (RMSE), mean absolute error (MAE) ~\cite{hodson2022root}, accuracy (ACC) and area under the receiver operating characteristic curve (AUC) \cite{ling2003auc}. Another possible reason to divide the training-testing set instead of calculating the goodness-of-fit on the whole dataset is that machine learning-based CDMs, especially deep learning-based CDMs, are more sophisticated and have much stronger fitting ability. If we focus on goodness-of-fit, then the diagnostic results could be easily overfitting.

\textit{\textbf{Explainability.}} Explainable diagnostic results are extremely important for providing feedback to users as well as downstream applications. The explainability of diagnostic results is not easy to define, and is still underexplored. Some studies use the metric called ``degree of agreement'' (DOA) to measure the explainability of diagnostic results~\cite{chen2017tracking,wang2020neural,chen2024disentangling}. The assumption behind DOA is that, if an examinee $a$ has a higher proficiency on knowledge concept $k$ than another examinee $b$, then $a$ is supposed to perform better than $b$ on test items requiring $k$. Wang et al.~\cite{wang2023dynamic} proposed partial DOA (PDOA) to mitigate the defect of DOA when dealing with test items requiring multiple knowledge concepts. Li et al.~\cite{li2024towards} used a similar but reverse metric called ``degree of consistency'' (DOC). In DOC, the assumption is that, if an examinee $a$ performs better than another examinee $b$ on test items requiring knowledge concept $k$, then the diagnosed $a$'s proficiency of $k$ should be higher than $b$'s proficiency of $k$. Without certain metrics, most works chose to analyze the explainability with diagnosed cases.

\textit{\textbf{Uncertainty/Confidence.}} Overall, the evaluation of the uncertainty or confidence for machine learning-based CDMs is still an underexplored topic. Here are a few studies. Bi et al.~\cite{bi2023beta} incorporated model uncertainty quantification into a meta-learned cognitive diagnosis framework by considering ability parameters and meta parameters as fully factorized Gaussian distributions, leading to lower expected calibration error (ECE). Ma et al.~\cite{zhang2023relicd} also used Gaussian distributions to represent ability parameters and proposed ReliCD, where the deviations were seen as the indicator of uncertainty/confidence. An ECE loss was added to the objective function to further decrease the ECE. Both of the above works seem to focus more on improving the diagnostic performance, while incorporating the estimated uncertainty as a useful by-product and validated by ECE. Wang et al.~\cite{wang2024unified} focused on the uncertainty estimation for CDMs, and provided a unified method called UCD. UCD also adopted the Bayesian-based method and can be used for both non-deep learning and deep learning-based CDMs. The uncertainty of diagnostic parameters (i.e., examinees' and items' parameters) was characterized through their posterior probability distributions, and the deviations were factorized into the data aspect and model aspect. The estimated uncertainty was validated using the PICP, PIAW metrics, and some statistical analysis.

\textbf{Discussion.} There exists quite a few differences between the evaluation of diagnostic results for Psychometrics-based CDMs and machine learning-based CDMs. Basically, the evaluation of accuracy for Psychometrics-based CDMs pays more attention to either accurately re-estimating the simulated parameters in synthesized datasets or better fitting the responses in real-world datasets. By contrast, the evaluation of accuracy for deep learning-based CDMs adopts a training-testing set division of real-world datasets to ensure the generality. Performance prediction task is mostly adopted to indirectly evaluate the accuracy. Moreover, the evaluation of cognitive diagnostic results should be comprehensive. The research about the evaluation of machine learning-based CDMs is still in the initial stage. 

In addition to the evaluation of CDMs, there also exist measurements of the reliability and validity of items within a test. The reliability focuses on whether the items receive consistent responses across different times and conditions through the test-retest method, Cronbach's alpha coefficient~\cite{tavakol2011making}, etc. The validity of items evaluates the extent to which the items can measure examinees' cognitive status. Validity can be assessed through various methods such as correlation analysis \cite{gogtay2017principles}, and factor analysis \cite{kline2014easy}.

\section{Applications}  \label{sec:application}
Since the inception of cognitive diagnosis, it has garnered widespread attention and has found applications in diverse fields. This section aims to provide a primary summary of its applications in intelligent education, covering areas such as computer adaptive testing and educational recommendation systems. Additionally, we will delve into the extended applications of cognitive diagnosis-related technologies in other domains.

\subsection{Applications in Education}
The most straightforward application of cognitive diagnosis in Education is to generate learning status diagnostic reports based on the diagnostic results, which help learners together with teachers to better understand the learning status of learners, and further serve as the basis of personalized applications, including the recommendation of learning resources and learning paths, and computerized adaptive testing. 

\textit{\textbf{Cognitive Diagnostic Reports.}}
Proverbially, both teachers (including intelligent tutoring systems) and learners need diagnostic reports on the cognitive status of learners. As for the teachers, they can use the diagnostic reports to check: (1) What are the learners' characteristics such as diligence, laziness, and inattention? (2) Whether or to what extent the learners have mastered a learning unit. As for the students, they need the diagnostic reports to check: (1) Whether or to what extent they have achieved the learning goal to adjust their learning styles and keep their learning enthusiasm~\cite{Goodman2004StudentTS}. To construct cognitive diagnostic reports, Roberts et al.~\cite{roberts2010developing} proposed a framework, that presents a graphical representation of the skill-level performance of individual students, to provide structured cognitive diagnostic reports with the Attribute Hierarchy Method. Zeniski et al.~\cite{zenisky2012developing} advanced a pipeline of diagnostic report development that is deﬁned by seven guiding principles for report design and validation. Maas et al.~\cite{maas2022cognitive} developed a personalized student dashboard, which provides a visual summary of student performance and outlines how this information can guide study behavior. Furthermore, online educational platforms (e.g., Zhixue~\footnote{https://www.zhixue.com/login.html}, Eedi~\footnote{https://family.eedi.com/}) have already developed various cognitive diagnostic reports, which include radar figures, learning paths, and other relevant panels or tables, to provide sufficient reports of cognitive status.

\textit{\textbf{Educational Recommendation Systems.}}
Typically, students pursue studies to meet specific learning objectives, such as mastering a particular knowledge domain or passing examinations successfully. Aligned with these predefined objectives, students necessitate tailored learning materials (typically exercises) to attain their desired outcomes. In conventional learning methods, there are two predominant strategies for resource selection, i.e., by students themselves and by professional teachers. Yet, the former strategy might lead to students selecting learning resources that are either overly basic or excessively advanced, consequently impeding optimal learning efficiency. Conversely, the latter method could potentially create higher barriers to access~\cite{desmarais2012review, ma2022exercise}. To overcome such problems, recent intelligent tutoring systems (e.g., ouc-online~\footnote{http://one.ouchn.cn/}) have utilized cognitive diagnosis methods, which estimate the cognitive state of each student, to choose the best appropriate learning resources based on their recommendation strategies and finally provide automatic educational recommendations for individual students.
In addition, Ma et al.~\cite{ma2022exercise} introduced the neutrosophic set method to compute the similarity between the cognitive states of students and recommends exercises. Chen et al.~\cite{cheng2021exercise} relied on the knowledge space theory to recommend tailored exercises based on estimated students' cognitive states and knowledge structures. Liu et al.~\cite{liu2019exploiting} utilized the cognitive diagnosis model to offer rewards to their reinforcement learning-based learning path recommending strategy.

\textit{\textbf{Computerized Adaptive Testing.}}
Traditionally, teachers hold a pencil-and-paper test to accomplish the student assessment by carefully selecting a fixed set of questions for all examinees at once. While this method effectively evaluates their performance and presents a uniform testing environment for all, it is challenging to ensure that the test items are properly selected for each examinee~\cite{bergstrom1992ability}. Consequently, recent efforts~\cite{ghosh2021bobcat, zhuang2023bounded} have shifted focus toward Computerized Adaptive Testing (CAT). CAT aims to provide tests that adapt dynamically to each examinee by tailoring test items based on the examinee's performance. It has several advantages, including heightened accuracy, shorter test length, enhanced security, and increased examinee engagement. CAT has already been successfully implemented by some standard test organizations~\cite{wendler2014research} like the Graduate Management Admission Test (GMAT)\footnote{The Graduate Management Admission Test (GMAT) stands as the most widely utilized test for admission into graduate business and management programs worldwide.} and the Graduate Record Examinations (GRE).

CDMs are the essential component of CAT, as the estimated ability levels ($\theta$) constitute the basis of item selection. For instance, \cite{chang1996global} employed Kullback-Leibler information, computed based on the unidimensional $\theta$ estimated by IRT, to assess item informativeness for each examinee and select the most informative one as the next to be assigned. Similarly, \cite{lord2012applications} used the estimated $\theta$ to compute the Fisher information measure for candidate items.
Recently, some novel methods based on advanced machine learning have been proposed. For example, Bi et al.~\cite{bi2020quality} integrated the concept from active learning, utilizing the model-agnostic expected model change derived from CDMs as the informative measure to select items. Additionally, \cite{ghosh2021bobcat, wang2023gmocat} formulated the CAT process as a Markov Decision Process (MDP) and utilized a reinforcement learning paradigm to address it. These approaches leverage CDMs to handle the state within their MDP formulation. The study by~\cite{zhuang2023bounded} innovatively transforms the CAT task into a coreset selection problem. This transformation involves aligning the gradients of the CDMs between a scenario with limited items and one with a complete set of items.

\subsection{Applications in Other Domains}

In recent years, some scholars have expanded the application of cognitive diagnosis-related technologies into other contexts, achieving notable success in the process.

\textit{\textbf{Truth Inference.}}
Deep learning generally relies on large-scale annotated data, often labeled through crowdsourcing (e.g., Amazon Mechanical Turk). Truth inference~\cite{whitehill2009whose}, in this context, refers to the technique of identifying the true data labels from potentially conflicting annotations made by annotators with varying abilities and backgrounds. In~\cite{whitehill2009whose, khattakaccurate} the authors utilized IRT-based probabilistic approaches to iteratively estimate the abilities of annotators and infer the true labels of images based on the annotations and abilities. Li et al.~\cite{li2016active} extended the previous IRT-based methods with compressive sensing theory and then mitigated human labeling errors. 

\textit{\textbf{Corporate Recruitment.}}
Assessing the skill qualifications of job seekers facilitates better matching between seekers and job requirements in online recruitment services~\cite{qin2018enhancing, zhu2018person}. The study in~\cite{qin2018enhancing} utilizes a word-level semantic representation module to represent job requirements and seekers' abilities, and then predict the compatibility with the hierarchical ability-aware attention strategies. Zhu et al.~\cite{zhu2018person} proposed an end-to-end convolutional neural network (CNN) to learn the joint representation of Person-Job fitness between requirements and abilities. Recent works like~\cite{qin2020enhanced, bian2020learning} proposed advanced neural networks, seeking more patterns to better mine job requirements and seekers' abilities, then compute the compatibility between them.

\textit{\textbf{Sport.}}
The assessment of specific skills is vital for athlete development~\cite{ando2018validity}. 
Matsuoka et al.~\cite{matsuoka2021development} introduced an IRT-based system comprising item construction, decision tree analysis, and test characteristic analysis. This system effectively evaluates defensive transitions in soccer games. Additionally, the research in \cite{yuzhou2018research} explored multi-directional training and technical analysis of basketball players using neural networks. By analyzing basketball players' abilities, this approach enables optimization of their training and team strategy. These assessment tools provide a tailored means for players, coaches, and managers to monitor the progression of individual skills throughout the training process.

\textit{\textbf{Game.}}
Precisely estimating players' abilities and properly arranging multiple players of comparable ability into competitive games, namely matchmaking, is an important component of online games~\cite{deng2021globally}. Its quality directly determines player satisfaction and further affects the life cycle of game products. The study in~\cite{gong2020optmatch} proposed a two-stage data-driven matchmaking framework, which firstly learns the low-dimensional abilities representations of individuals by capturing the high-order inter-personal interactions and then incorporates the team-up effect and predicts the match outcomes. The study in~\cite{wang2022graph} modeled the players and their win-loss relationships as an undirected weighted skill gap graph. By matching players properly after estimating their abilities, these works provide players with a considerable gain in their game experience.

\textit{\textbf{Psychological and Physical Health Diagnosis.}}
As the technique arose from psychometrics, it is natural to use CDMs to diagnose patients' mental health. For instance, Fraley et al.~\cite{fraley2000item} employed item response theory to diagnose the existence of adult attachment from self-report measures. Templin et al.~\cite{templin2006measurement} utilized the DINO model to assess and diagnose pathological gamblers. Tu et al.~\cite{tu2017new} designed questions related to internet addiction and further utilized G-DINA to diagnose whether a subject has internet addiction. In addition, CDMs are also used in diagnosing physical health. Liang et al.~\cite{liang2023measuring} employed G-DINA, in conjunction with constructed data and a Q-matrix, to predict six-month Quality of Life (QoL) in breast cancer.
The results from CDMs provide valuable references for doctors to assess examinees' health conditions and support planning the treatments.

\section{Datasets and CDM tools} \label{sec:datasetTool}
\subsection{Datasets}
To further help researchers who have an interest in developing CDMs, we summarize some frequently used datasets in the related works. As the datasets used in early research papers, i.e., psychometrics-based CDMs, are mostly synthesized or unavailable, here we only summarize publicly available datasets from recent research papers. Moreover, we have open-sourced a Python library called EduData\footnote{\url{https://github.com/bigdata-ustc/EduData/}} which provides easy access to numerous datasets.

Basically, the datasets used in CDM research can be classified into two types, i.e., from standard tests and from e-learning platforms. Table \ref{tab:dataset} presents some basic descriptions including statistics of the datasets.
\begin{table}[tb]
    \centering
    \caption{Basic descriptions of the datasets.}
    \begin{tabular}{cc|ccccc}
        \toprule
        Source & Dataset & \#Examinee & \#Item & \#Response & \#KC/Skill & Extra info \\
        \midrule
        \multirow{5}{1.4cm}{Standard tests} & FrcSub & 536 & 20 & 10,720 & 8 & \ding{55}  \\
        & Math1 & 4,209 & 20 & 84,180 & 11 & \ding{55}  \\
        & Math2 & 3,911 & 20 & 78,220 & 16 & \ding{55}  \\
        & ECPE & 2,922 & 28 & 81,816 & 3 & \ding{55}  \\
        & PISA2015 & 519,334 & 183 & 12,612,424 & - & \ding{51}  \\
        \midrule
        \multirow{4}{1.4cm}{E-learning systems} & ASSISTments2009 & 4,163 & 17,751 & 346,860 & 123 & \ding{51}  \\
        & ASSISTments2012 & 46,674 & 179,999 & 6,123,270 & 265 & \ding{51}  \\
        & Junyi & 247,606 & 722 & 25,925,922 & 41 & \ding{51}  \\
        & Eedi & 118,971 & 27,613 & 15,867,850 & 288 & \ding{51}  \\
        \bottomrule
    \end{tabular}
    \label{tab:dataset}
\end{table}

\textbf{Datasets from standard tests.} This type of data is collected from standard tests. Therefore, In each dataset, all examinees have provided their responses to all test items. The data size is mostly small, with fewer examinees, test items, and knowledge concepts.
\begin{itemize}
    \item \textbf{FrcSub.} The FrcSub~\cite{tatsuoka1984analysis,liu2018fuzzy} is composed of the scores of middle school students on fraction subtraction objective problems. 
    \item \textbf{Math1 \& Math2.} The Math1 and Math2\footnote{\url{http://staff.ustc.edu.cn/~qiliuql/data/math2015.rar}} datasets are collected from two final mathematical exams from high school students, including both objective and subjective problems.
    \item \textbf{ECPE.} The full name of this dataset is \textit{Examination for the Certificate of Proficiency in English}\footnote{\url{https://rdrr.io/cran/GDINA/man/ecpe.html}}. It is collected from a standard English test by the English Language Institute of the University of Michigan and is well-adopted in educational psychology.
    \item \textbf{PISA2015.} The PISA2015\footnote{\url{http://www.oecd.org/pisa/data/2015database/}} dataset is released by OECD's Programme for International Student Assessment. PISA is a worldwide testing program that measures 15-year-olds' abilities to address real-life challenges. Four core domains of PISA2015 include science, reading, mathematics and collaborative problem-solving. In addition, PISA also collects students' background information such as region, home economic and cultural status through questionnaires. The test is put out every three years, and PISA2015 is the result released in 2015. Not every student answers every question as many versions of the computer exam exist. The assessed abilities are not barely the mastery of knowledge concepts and vary in different core domains. For example, in the science domain, the assessment tasks focused on \textit{Competencies}, \textit{Knowledge}, and \textit{Context}. The \textit{Knowledge} further includes \textit{Content Knowledge}, \textit{Procedural Knowledge} and \textit{Epistemic Knowledge}. The usage of these types of abilities is better decided by researchers and we choose not to simply put their count in the table.
\end{itemize}

\textbf{Datasets from e-learning systems.} This type of data is collected from e-learning systems, especially online learning platforms. The responses of a learner may be distributed over either a short or a long period of time, which might be considered as a violation of the assumption mentioned in the Overview, i.e., constant ability. The requirements for the dataset are sometimes not very strict, or the researchers have already assumed that students' abilities are relatively stable within the datasets. Some researchers have mentioned this problem and made analyses and preprocesses. For instance, analyzing the stability of learners' abilities~\cite{wang2020neural}, constructing a subset of the original dataset where the responses are collected in a shorter period of time~\cite{li2022hiercdf,wang2024unified}, and only maintain the first response to an item when a learner has multiple attempts on it.
\begin{itemize}
    \item \textbf{ASSISTments2009 \& ASSISTments2012.}\footnote{\url{https://sites.google.com/site/assistmentsdata}} Both of these two datasets are collected from ASSISTments, an online tutoring system in the United States. The full name of ASSISTments2009
    is the ASSISTments 2009-2010 skill builder data set. ASSISTments2009 is collected during the school year from 2009 to 2010, and students were asked to practice the questions related to similar knowledge concepts until they answered three or more times in a row. It is worth noting that there are multiple versions of ASSISTments2009. Early versions have several problems that may have caused some unreliable experiments in early research papers~\cite{xiong2016going}. The final version has solved the problems.
    ASSISTments2012 is collected during the school year from 2012 to 2013. ASSISTments2012 contains more students and responses. However, the majority of the test items are not labeled to any knowledge concepts, and each test item is labeled to no more than one related knowledge concept. Researchers can request access to the item contents by sending emails to the providers (see the website for details).
    Both ASSISTments2009 and ASSISTments2012 provide some side information, such as the attempt counts, start time, end time, and problem type. 
    \item \textbf{Junyi.}\footnote{\url{http://www.junyiacademy.org/}} The Junyi dataset contains student online learning logs on mathematical exercises which are collected from a Chinese online learning platform called Junyi Academy. The dataset contains practicing logs from Oct. 2012 to Jan. 2015, exercise-related information on the platform, and annotations of exercise relationships.
    \item \textbf{Eedi.} Eedi is the dataset released by the NeurIPS 2020 education challenge\footnote{\url{https://competitions.codalab.org/competitions/25449}}, containing students' answers to mathematics questions from Eedi, an online educational platform. All items are 4-choice questions with only one correct choice. In addition, Eedi also provides side information such as gender, date of birth, group ID, and quiz ID.
\end{itemize}

\subsection{CDM Tools}
The implementations of CDMs are not standard and are mostly developed independently by researchers. In earlier research, psychometrics-based CDMs were usually implemented with R language (a programming language suitable for statistical analysis). Most of their codes were not properly shared and are difficult to access now. Additionally, thanks to the relatively longer and more mature research on psychometrics-based CDMs, there are already some related software and platforms. For example, the Vector Psychometric Group\footnote{\url{https://vpgcentral.com/software/}} has launched several useful software for cognitive diagnosis, such as Adaptest, flexMIRT, and IRTPRO.
Tu et al.~\cite{tu2023flexcdms} launched a web-based cognitive diagnosis platform called flexCDMs\footnote{\url{http://www.psychometrics-studio.cn/}}, which provides easy usage of traditional DCMs such as DINA, DINA, rRUM, and GDM. 

The research for deep learning-based CDMs prefers Python to implement their models. Although the code availability is better, it is still laborious to search for the CDMs and learn about the codes with different programming styles. Therefore, we have developed an open-sourced Python library called EduCDM\footnote{\url{https://github.com/bigdata-ustc/EduCDM}}, which provides easy use of numerous CDMs. The code structure of EduCDM is more unified and thus easier to understand, modify and extend. We will keep updating this library.

\section{Discussion of Future Research Directions} \label{sec:future}
Despite the promising performances achieved by the existing state-of-the-art CDMs, limits as well as opportunities exist that encourage future research.

\textit{\textbf{Cognitive diagnosis in more application areas.}}
Cognitive diagnosis has obtained many achievements in traditional application areas, especially in intelligent education and psychological measurement. However, the demand for individual ability evaluation is not limited to the traditional areas. For instance, evaluating the proficiency of trial lawyers in different legal fields enables matching qualified lawyers to strive for the clients' best rights while ensuring fairness and litigation \cite{an2021lawyerpan}. In companies, effective assessment of employees' abilities helps with enhancing employee training and development \cite{rodriguez2017importance,wang2020personalized}. Moreover, the objects of diagnosis can be AI agents to assess their anthropomorphic intelligence. Zhuang et al. \cite{zhuang2023efficiently} made an attempt to measure the cognitive ability of large language models (LLM) through cognitive diagnosis and adaptive testing. How to measure the cognitive ability of large models such as LLM and how to leverage large models for better diagnosing human ability are worthy study. It is expected that cognitive diagnosis can be developed for wider areas and applications.

\textit{\textbf{More question types and multimodal data.}}
Most cognitive diagnosis models handle binary responses, which belong to questions having only ``correct'' or ``incorrect'' answers. However, some types of questions can have partially correct answers. For instance, cognitive diagnosis using polytomous responses has caught some attention in traditional studies \cite{Masters1982pcm,Andersen1997ratingsm}, while it is still underexplored with deep-learning-based modeling. Another example is the questions requiring more extensive text responses, such as writing and programming, which have a more complex scoring structure and are also not sufficiently studied. In terms of data types, existing models mainly consider information such as question text \cite{wang2022neuralcd}, knowledge concept relations \cite{gao2021rcd}, participant background features \cite{zhou2021modeling}, etc. The behavior of participants is considered by some works in a coarse manner, such as the number of attempts, hints, and response time \cite{wu2017knowledge,thai2015multi,zhan2018cognitive,zhan2022cognitive}. For questions such as writing and programming, fine-grained answering progress (e.g., type in, delete, copy and paste) is underexplored. The primary obstacle on this path is likely to be the collection and disclosure of data.

\textit{\textbf{More than the model structure.}}
While the concentration of this survey is the cognitive diagnosis models, especially the recent development of deep learning-based models and their various applications, we would like to point out some issues besides the model structure designing that are ignored by recent research. 1) The definitions of knowledge concepts (or skills) and their structures. In the field of education, such definitions typically come from experts' professional knowledge and rigorous discussion.
However, the definition of knowledge concepts in other application areas can be less rigorous and systematic. 2) Calibration of item parameters. In a more standard cognitive diagnosis process, there is a calibration stage during which responses to the candidate items are provided by a certain group of examinees. The examinees are supposed to have ability levels following a certain distribution (e.g., normal distribution \cite{ackerman1989unidimensional,kim2006comparative}). After that, item parameters are estimated based on these responses. Suitable items are then selected into the item bank and construct the test papers \cite{reckase2010designing}. For the true test takers who are assigned to these items, their ability parameters are estimated. In the recent deep learning-based methods, calibration of item parameters is usually omitted or simplified to facilitate the usage, where the parameters of items and test takers are estimated simultaneously, with less consideration of the influence of test takers' ability distribution upon the parameter estimation. 3) Comparability of diagnostic results. For continuous CDMs, the diagnostic results, i.e., the estimated parameters of test takers, are most meaningful within the trained model using the corresponding response data.
However, the diagnostic results among different trained CDMs, even having the same model structure trained with different data, are not directly comparable. 
As a measurement tool, the comparability of diagnostic results across different models is of great use in circumstances such as ability comparison among test takers across time/institutions, and adding new items into a CAT item bank. Related works called ``parameter linking'' have been proposed for CDMs with simple structures including IRT and MIRT \cite{Reckase2009,van2016linking,kang2012linking}. However, for state-of-the-art CDMs, which are more complicated, the comparability problem is still underexplored. 4) Cognitive diagnosis, if cooperated with some interpretative methods, can be used for discovering examinees' cognitive patterns from empirical data in a data-driven way. Liu et al.~\cite{liu2024automated} have pioneered in this direction and proved its feasibility. 

\textit{\textbf{Model evaluation.}} Model evaluation is an important yet easily neglected problem of cognitive diagnosis.  Different from most machine learning models focusing on accurately predicting something either by classification or regression, cognitive diagnosis pays more attention on the diagnostic results. Although the student performance prediction task is adopted to indirectly validate the accuracy of diagnostic results due to the lack of ground truth of examinees' abilities, excessive focus on the task of student performance prediction will gradually turn the cognitive diagnostic model into a predictive model rather than a diagnostic model. The evaluation of cognitive diagnosis models should be multi-faceted. Especially, the research on evaluation metrics for SOTA deep learning-based CDMs is still in its infancy and relatively lacking. We hope there will be insightful research on CDM evaluation from different aspects, including but not limited to accuracy, explainability, uncertainty, identifiability~\cite{li2024towards}, and stability of parameter estimation.

\section{Conclusion}  \label{sec:conclusion}
In this survey, we have reviewed the development of models for cognitive diagnosis. Basically, the research history is divided into two stages: psychometrics-based CDMs and machine learning-based CDMs, between which the latter is the key emphasis of this survey. Through reviewing and comparing existing research outcomes, we have found that the transition from psychometrics-based CDMs to machine learning-based CDMs has undergone changes in not only model structures but also data types. Moreover, the research topics become more diverse. The trend in model structure is the shift from psychometrics methods where the interaction functions are designed by experts to data-driven deep learning methods. In terms of data types, behavioral data from online learning is gradually being taken into consideration, and the role of text, images, graphs, and other types of data in cognitive diagnosis is being explored. Furthermore, researchers are also beginning to consider issues such as cold start and fairness in cognitive diagnosis. Besides these, we have also summarized some main changes in the parameter estimation and model evaluation approaches and provided some examples of where cognitive diagnosis has been applied. 

We hope this survey can invoke more attention on cognitive diagnosis, and inspire more interesting and insightful research in this area. We have summarized some commonly used datasets and useful tools for the application and research of cognitive diagnosis. In addition, we have also discussed several promising future research directions. In summary, we advocate for the further development of cognitive diagnostic methods in more fields and from more diversified perspectives. However, we also remind that attention needs to be paid to the fact that cognitive diagnosis models are measurement tools for human ability, and avoid placing too much emphasis on predicting learners' question-answering performance.

We also recognize some limitations of this survey. First, since this survey focuses more on the recent progress of cognitive diagnosis, i.e., deep learning-based CDMs, the summary of the traditional psychometrics-based works may not be comprehensive enough. The related work of traditional methods is rich and relatively mature thanks to decades of research~\cite{dibello200631a,leighton2007cognitive,reckase2010designing}. Readers can refer to relevant literature to gain a deeper understanding. Second, as new research work on cognitive diagnosis continues to emerge and is scattered across different publications in the fields of education and computer science, we may have missed a few noteworthy works. If necessary, we will continue to track the progress in this direction in future discussions.

\bibliographystyle{ACM-Reference-Format}
\bibliography{reference}



\end{document}